\documentclass[runningheads]{llncs}
\usepackage{graphicx}

\usepackage{tikz}
\usepackage{comment}
\usepackage{amsmath,amssymb} 
\usepackage{color}

\usepackage[accsupp]{axessibility}  

\usepackage[utf8]{inputenc} 
\usepackage[T1]{fontenc}    
\usepackage[pagebackref,breaklinks,colorlinks,citecolor=green,linkcolor=red,urlcolor=blue]{hyperref}       
\usepackage{url}            
\usepackage{booktabs}       
\usepackage{amsfonts}       
\usepackage{nicefrac}       
\usepackage{microtype}      
\usepackage{colortbl}
\usepackage{xcolor}         
\usepackage{amsmath}
\usepackage{bm}
\usepackage{amssymb}            
\usepackage{multirow}
\usepackage[thicklines]{cancel}
\usepackage{caption}
\usepackage{subfigure}
\usepackage{wrapfig}
\usepackage{algorithm}
\usepackage[noend]{algpseudocode}
\usepackage{algorithmicx}
\usepackage{float}
\usepackage{placeins}
\usepackage{bookmark}

\definecolor{green}{HTML}{00aa00}
\definecolor{blue}{HTML}{0000aa}
\definecolor{red}{HTML}{aa0000}
\definecolor{salmon}{HTML}{fa8072}

\definecolor{bg-ssad}{HTML}{fae9df}
\definecolor{bg-anoonly}{HTML}{e4eef4}

\newcommand{\trm} {\textrm}

\newcommand{\tbf} {\textbf}

\newcommand{\mcal}{\mathcal}

\newcommand{\ssubsec}{\vspace{0pt}\tbf}

\newcommand{\diffH}[1]{\textcolor{blue}{\rlap{$_{\uparrow #1}$}} }
\newcommand{\diffL}[1]{\textcolor{red}{\rlap{$_{\downarrow #1}$}}}

\begin{document}
\pagestyle{headings}
\mainmatter

\title{AnoOnly: Semi-Supervised Anomaly Detection with the Only Loss on Anomalies}

\author{
  Yixuan Zhou\inst{1}\orcidID{0000-0003-1397-9396} \and
  Peiyu Yang\inst{1} \and
  Yi Qu\inst{1} \and
  Xing Xu\inst{1} \and
  Zhe Sun\inst{2} \and
  Andrzej Cichocki\inst{3}\orcidID{0000-0002-8364-7226}
  }
\authorrunning{Y. Zhou et al.}
\institute{
  Center for Future Media \& School of Computer Science and Engineering, University of Electronic Science and Technology of China, China \and
  Juntendo University, Japan \and
  RIKEN, Japan
}

\maketitle

\begin{abstract}
  Semi-supervised anomaly detection (SSAD) methods have demonstrated their effectiveness in enhancing unsupervised anomaly detection (UAD) by leveraging few-shot but instructive abnormal instances.
  However, the dominance of homogeneous normal data over anomalies biases the SSAD models against effectively perceiving anomalies.
  To address this issue and achieve balanced supervision between heavily imbalanced normal and abnormal data, we develop a novel framework called \emph{AnoOnly} (\tbf{Ano}maly \tbf{Only}).
  Unlike existing SSAD methods that resort to strict loss supervision, AnoOnly suspends it and introduces a form of weak supervision for normal data.
  This weak supervision is instantiated through the utilization of batch normalization, which implicitly performs cluster learning on normal data.
  When integrated into existing SSAD methods, the proposed AnoOnly demonstrates remarkable performance enhancements across various models and datasets, achieving new state-of-the-art performance. 
  Additionally, our AnoOnly is natively robust to label noise when suffering from data contamination.
  Our code is publicly available at \url{https://github.com/cool-xuan/AnoOnly}.
\end{abstract}

\section{Introduciton}
Anomaly Detection (AD), also known as outlier detection \cite{chalapathy2019deep,chandola2009anomaly,cevikalp2023anomaly}, is an already-intensely-studied task and drives practical applications in various domains such as medical diagnosis \cite{tschuchnig2022anomaly}, industrial inspection\cite{you2022unified,zavrtanik2021reconstruction}.
Over the past few decades, plentiful AD algorithms have been proposed with the majority of them developed in an unsupervised fashion \cite{ergen2019unsupervised,nicolau2018learning} (referred to as UAD), where only normal data is available for training. Most UAD methods primarily focus on representation learning to capture the inherent characteristics of normal data \cite{ruff2018deep,zong2018deep}.
Some recent studies \cite{han2022adbench,jiang2023weakly} consider collecting and labeling a handful of abnormal samples to be feasible, emerging several semi-supervised anomaly detection (SSAD) methods \cite{akcay2019ganomaly,kumagai2021semi,ruff2019deep,zhao2018xgbod}.
Through leveraging limited yet informative abnormal data, SSAD methods effectively facilitate perceptual capabilities and robustness towards anomalies \cite{han2022adbench}.

Existing semi-supervised anomaly detection methods \cite{akcay2019ganomaly,zhao2018xgbod,pang2018learning,pang2019devnet,zhou2021feature} commonly adopt a straightforward approach: they integrate additional loss supervision towards partial abnormal instances on top of loss functions designed in UAD methods \cite{ruff2018deep}, as illustrated in Fig. \ref{fig:framework-ssad}.
The underlying principle of these SSAD methods is to concurrently encourage normal instances to align closely with a predefined center and push anomalous instances as far as possible from this center.
Nevertheless, the significant data distribution imbalance between abundant normal data and limited abnormal data often leads to a bias towards the dominant normal data in the trained SSAD models \cite{yang2020rethinking}, thereby lacking sufficient emphasis on learning the discriminative characteristics of anomalies.
Although over-sampling is employed by prior methods \cite{pang2019devnet,pang2019deep} to rebalance supervision, it inevitably introduces the risk of overfitting to a limited set of duplicated abnormal instances \cite{hawkins2004problem}.

\begin{figure*}[t]
  \centering
  \begin{minipage}[c]{\linewidth}
    \centering
    \includegraphics[width=\linewidth]{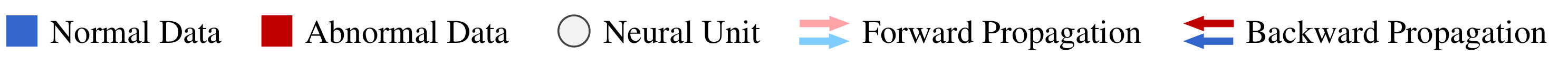}
  \end{minipage}

  \setcounter{figure}{1}
  {\begin{minipage}[b]{0.33\linewidth}
      \centering
      \includegraphics[width=\linewidth]{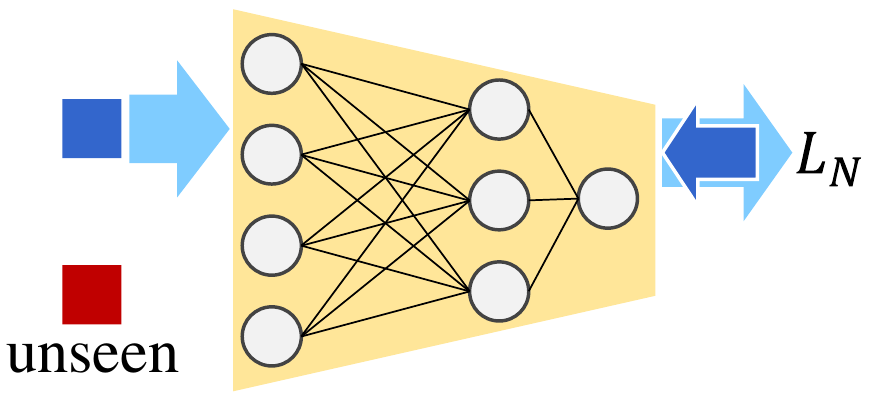}
      \captionof{subfigure}{UAD methods \quad\qquad} \label{fig:framework-uad}
    \end{minipage}
    \hfill
    \begin{minipage}[b]{0.31\linewidth}
      \centering
      \includegraphics[width=\linewidth]{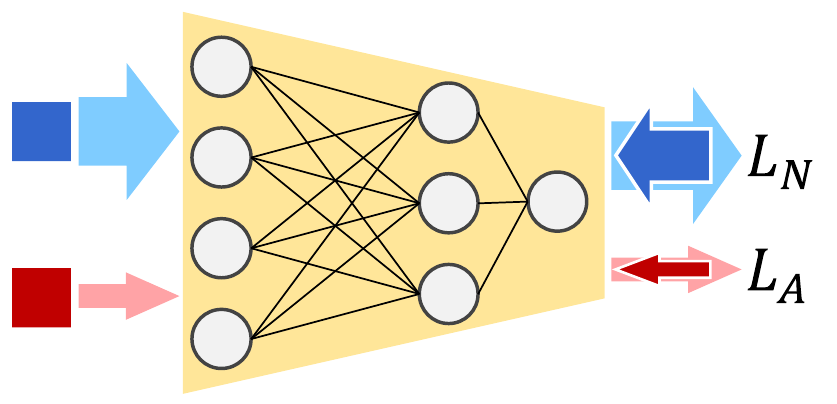}
      \captionof{subfigure}{SSAD methods\quad\qquad} \label{fig:framework-ssad}
    \end{minipage}
    \hfill
    \begin{minipage}[b]{0.31\linewidth}
      \centering
      \includegraphics[width=\linewidth]{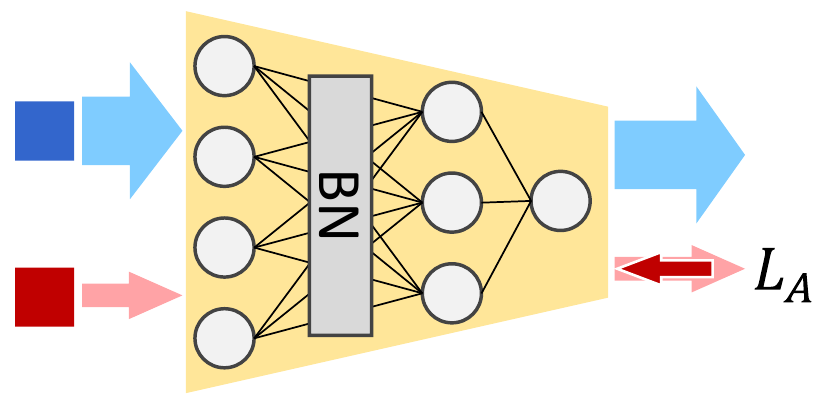}
      \captionof{subfigure}{Our AnoOnly \qquad\qquad} \label{fig:framework-anoonly}
    \end{minipage}}
  \setcounter{figure}{0}
  \caption{Illustration of UAD methods, SSAD methods and our AnoOnly. The training data amount of normal and anomaly is visualized by the thickness of the arrow. Our AnoOnly omits the loss supervision on normal data yet introduces BN to implicitly perform online cluster learning, which can be deemed as a form of weak supervision.} \label{fig:framework}
\end{figure*}

To mitigate supervision imbalance and enhance the discriminative capability targeting anomalies, we propose a peculiar SSAD framework named \emph{AnoOnly} (\tbf{Ano}maly \tbf{Only}) as demonstrated in Fig. \ref{fig:framework-anoonly}, which retains the strict loss supervision on few-shot anomalies while suspending that for abundant normal data.
Despite the removal of rigorous loss supervision, we introduce batch normalization \cite{ioffe2015batch} (BN) as a form of weak supervision to capture statistical characteristics \cite{kirichenko2020normalizing} within the numerous normal data.
The strategic utilization of BN in our AnoOnly effectively mitigates the training bias arising from the dominant presence of normal data over anomalies.
By explicitly redirecting the trained model towards anomaly discrimination, AnoOnly achieves a more equitable and effective semi-supervised training approach.
Particularly, the proposed AnoOnly framework is compatible with existing SSAD methods \cite{ruff2019deep,pang2019devnet,pang2019deep,zhou2021feature}, yielding a remarkable improvement in detection performance without any sacrifice or burden. We summarize our contributions as follows:

\begin{itemize}
  \item We introduce a novel SSAD framework called \emph{AnoOnly} to address the supervision bias to ample normal data. The proposed AnoOnly replaces strict loss supervision on normal data with a form of weak supervision, effectively rebalancing the supervision towards anomalies.
  \item In our AnoOnly framework, despite the absence of explicit loss supervision on normal data, we disclose that batch normalization inherently introduces a form of weak supervision for normal data by capturing statistical characteristics. Furthermore, we identify the weak supervision provided by batch normalization to be online cluster learning.
  \item Our AnoOnly is plug-and-play and significantly enhances the performance of prior SSAD methods across diverse datasets, achieving a new state-of-the-art performance. Our AnoOnly also demonstrates strong generalization to unseen anomalous types and robustness to label noise, which is practical for real-world scenarios.
\end{itemize}

\section{Related works}

Anomaly detection plays a critical role in broader applications such as medical diagnosis \cite{tschuchnig2022anomaly}, industrial inspection \cite{zhou2020variational}, social media analysis \cite{cauteruccio2021framework,zhao2020multi}, and video surveillance \cite{chen2022mgfn,feng2021mist,zaheer2022generative,wang2023memory}.
Recent advancements \cite{ruff2019deep,zhao2018xgbod,pang2019devnet,zhou2021feature} in Semi-Supervised Anomaly Detection (SSAD) have demonstrated their superiority over fully supervised \cite{gorishniy2021revisiting,kawachi2018complementary} or unsupervised \cite{ruff2018deep,zaheer2022generative,goldstein2016comparative,xu2018unsupervised,zhou2017anomaly} competitors by effectively leveraging limited abnormal data on the top of representation learning on sufficient normal data \cite{han2022adbench,pang2019deep,feng2021mist,wu2022self}.
However, due to the heavily imbalanced data distribution, models are prone to bias towards normal data, resulting in inferior anomaly detection performance \cite{akcay2019ganomaly,ruff2019deep,pang2018learning}.
To address this challenge, some recent studies \cite{pang2019devnet,zhou2021feature,pang2019deep} have attempted to augment supervision volume on anomalies by over-sampling strategies \cite{buda2018systematic,feng2021exploring}.
While over-sampling is a feasible approach to shift the model bias back to anomalies, it introduces a new challenge of overexposing limited instances of seen anomalies. The overexposure leads to overfitting of the model \cite{hawkins2004problem} to these specific anomalies, resulting in the reduced capability to detect anomalies of unseen types.

In prior works \cite{ruff2019deep,pang2019devnet}, cluster learning \cite{bo2020structural,caron2018deep,shah2018deep} has emerged as a potential solution to incorporate a weaker form of supervision on normal samples and mitigate supervision discrepancy over restricted anomalies.
For example, DeepSAD \cite{ruff2019deep}, a state-of-the-art semi-supervised approach, performs explicit cluster learning in the feature space learned by autoencoders \cite{masci2011dcae,radford2015dcgan}, followed by Support Vector Data Description (SVDD) \cite{tax2004svdd} to distinguish normal and abnormal instances.
However, the cluster center estimated by autoencoders remains fixed during SVDD training, serving as a constant vector with no contribution to the optimization of SVDD.
In contrast, we utilize BN \cite{ioffe2015batch} to capture statistical characteristics predominantly relied on normal data and implicitly perform online cluster learning alongside the only loss supervision of anomaly.
By taking full advantage of this online cluster learning mechanism, our AnoOnly framework effectively achieves supervision rebalance, yielding impressive improvements when seamlessly integrated with prior methods.

\section{Methodology}

\subsection{Preliminaries} \label{sec:pre}
We define semi-supervised anomaly detection (SSAD) as specifically referring to the task of detecting anomalies with incomplete labeling following prior works \cite{han2022adbench,ruff2019deep,pang2019devnet}.
In this case, the training dataset $\mathcal{D}$ is divided into a labeled dataset $\mathcal{D}_L = \{\bm{x}_{1},\dots,\bm{x}_{n}\}$ with limited labeled abnormal instances, alongside an unlabeled dataset $\mathcal{D}_U = \{\bm{x}_{n+1},\dots,\bm{x}_{n+m}\}$ containing a majority of normal data mixed with some unlabeled abnormal data, where $\bm{x} \in \bm{\mathcal{X}} \subseteq \mathbb{R}^d$.
Notice that the data amount of $\mathcal{D}_L$ is much smaller than that of $\mathcal{D}_U$, i.e. $n \ll m$.
Specifically, during the training process, we assign the label of $-1$ to labeled anomalies ($\mathcal{D}_L$) and the label of $+1$ to the instances within the unlabeled dataset $\mathcal{D}_U$, under the assumption that most of them are normal.
We justify this approximation on the grounds that most instances are normal due to the rarity of anomalies. Throughout this paper, unless explicitly emphasized, \emph{normal data} and \emph{unlabeled data} are considered equivalent.

Correspondingly, the aim of SSAD can be specified as training a neural network $\mathcal{F}(\cdot; \bm{\theta})$ with parameters $\bm{\theta}$ to estimate the abnormality of input data $\bm{x}$ through an anomaly scoring mechanism:
$\bm{\mathcal{X}} \xrightarrow{\mathcal{F}(\cdot; \bm{\theta})} \mathcal{S} \in \mathbb{R}$.
Although the specific loss functions vary among different methods \cite{ruff2019deep,zhao2018xgbod,zhou2021feature}, their objective function with weight regularization $R(\bm{\theta})$ can be uniformly formulated as follows:
\begin{equation}
  L(\mathcal{D};\theta)=L_N(\mathcal{D}_U;\bm{\theta})+L_A(\mathcal{D}_L;\bm{\theta})+\lambda R(\bm{\theta}),
\end{equation}
where $L_N$ encourages the anomaly score of unlabeled data $\mathcal{D}_U$ (treated as normal for training) to converge towards $0$, while $L_A$ simultaneously pushes the scores of labeled anomalies ($\mathcal{D}_L$) to be as large as possible.
For instance, the specific loss function $L_A$ targeting labeled abnormal instances is designed to encourage the anomaly score of anomalies to approach infinity in DeepSAD \cite{ruff2019deep} as follows:
\begin{align}
   & \quad L_N(\mathcal{D}_U;\bm{\theta}) \qquad\qquad L_A(\mathcal{D}_L;\bm{\theta}) \nonumber \\[-5pt]
  L^\trm{DeepSAD}(\mathcal{D};\theta)= \
   & \overbrace{\lambda_n \sum_{\mcal{D}_U} \left\Vert \mathcal{F}(\bm{x}; \bm{\theta}) \right\Vert ^2} + \
  \overbrace{\sum_{\mcal{D}_L} \left( \left\Vert \mathcal{F}(\bm{x}; \bm{\theta}) \right\Vert ^2 \right) ^ {-1}}
  +\lambda R(\bm{\theta}),
\end{align}
where $\lambda_n$ is the normal loss weight and set to 1 following the official implementation \cite{ruff2019deep}.

\begin{figure*}[t]
  \centering
  \includegraphics[width=\linewidth]{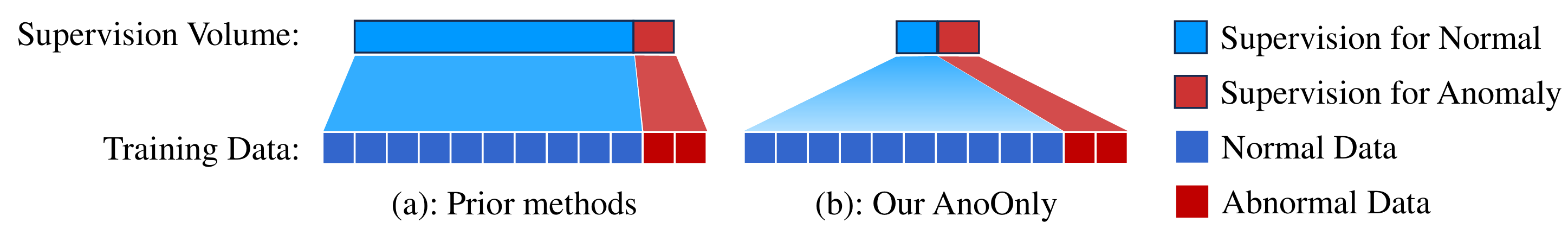}
  \caption{An intuitive illustration of the supervision volume for imbalanced normal and abnormal data of prior SSAD methods and our AnoOnly. Our AnoOnly replaces the strict loss supervision of normal with weak supervision, achieving equitable training on normal and anomaly.}
  \label{fig:supervision-volume}
\end{figure*}

\subsection{The proposed \emph{AnoOnly} framework}
As described in the preliminaries, prior SSAD methods \cite{akcay2019ganomaly,pang2018learning} train the model $\mathcal{F}(\cdot; \bm{\theta})$ by the loss functions involving both restricted abnormal data and sufficient normal data.
However, despite the abundance of normal training instances, they exhibit homogeneous characteristics, imparting deficient guidance for effectively discriminating diverse anomalies.
Additionally, the scarcity of abnormal training data relative to normal data leads to a muffled volume of loss supervision targeting a minority of anomalies as shown in Fig. \ref{fig:supervision-volume}\textcolor{red}{(a)}.
The discrepancy in training data amount raises a problem where the instructive information conveyed by the abnormal instances is overshadowed by numerous but monotonous characteristics of normal data.

To fully utilize the partially labeled abnormal data, we introduce a novel semi-supervised anomaly detection framework called AnoOnly to rebalance the supervision volume between the heavily imbalanced normal and abnormal data.
In contrast to the previous methods \cite{pang2019devnet,zhou2021feature,pang2019deep} that resort to the over-sampling strategy to amplify the supervision volume towards anomaly, our AnoOnly inversely weakens that for normal data.
As shown in Fig. \ref{fig:framework-anoonly}, the proposed framework only relies on loss supervision $L_A$ for labeled anomalies while eliminating $L_N$ for unlabeled data, thus naming it as \emph{AnoOnly} (\tbf{Ano}maly \tbf{Only}).
Specifically, our AnoOnly re-writes the modularized loss function of existing SSAD methods as follows:
\begin{equation} \label{eq:anoonly-loss}
  L^\trm{AnoOnly}(\mathcal{D};\theta)=\textcolor{gray!50!white}{\bcancel{L_N(\mathcal{D}_U;\bm{\theta})}+}L_A(\mathcal{D}_L;\bm{\theta})+\lambda R(\bm{\theta}).
\end{equation}

Despite the absence of strict loss supervision, normal data remains integral to our AnoOnly and provides weak supervision for model training.
This weak supervision is introduced through the utilization of batch normalization, which implicitly assumes the role of online cluster learning, as further elaborated in Section \ref{sec:bn}. 
Therefore, as illustrated in Fig. \ref{fig:supervision-volume}\textcolor{red}{(b)}, the proposed AnoOnly achieves a rebalancing of supervision volume by incorporating strong loss supervision on limited abnormal instances and weak supervision in the form of online cluster learning on abundant normal data.
Distinct from conventional SSAD models that typically exhibit a bias towards normal data, our AnoOnly effectively transfers this bias to enhanced anomaly perception.


\subsection{Weak supervision on normal data via batch normalization} \label{sec:bn}

\begin{wrapfigure}{R}{0.55\linewidth}
  \centering
  \includegraphics[width=1.0\linewidth]{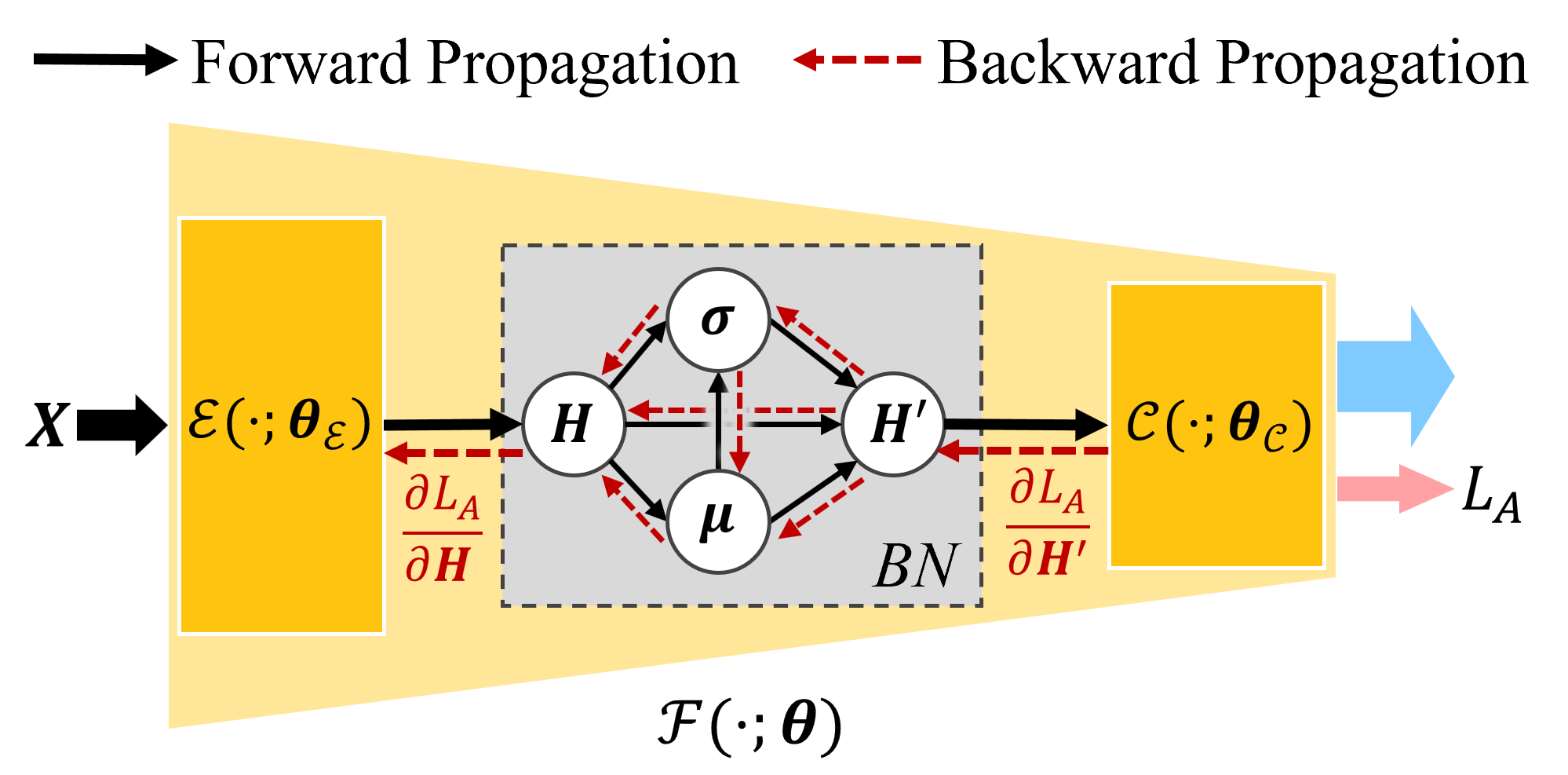}
  \caption{The forward and backward propagation of BN in $\mathcal{F}(\cdot; \bm{\theta})$.}
  \label{fig:bn}
\end{wrapfigure}

In our AnoOnly framework, besides stabilizing training, batch normalization (BN) also serves as implicit cluster learning to introduce weak supervision derived from normal data.
We first provide a notation definition before our illustration about the efficacy of BN in our AnoOnly design.
During training, the entire dataset $\mathcal{D}$ is randomly shuffled and divided into $k$ mini-batches $\{\mathcal{B}_i\}_{i=1}^k$ with the batch size of $b$, where $\mathcal{B}_1 \cup  ... \cup \mathcal{B}_k = \mathcal{D}$.
For each mini-batch $\mathcal{B}$, $X_\mcal{B} = \{\bm{x}_i\}_{i=1}^b$ refers to the batched inputs.
As illustrated in Fig. \ref{fig:bn}, the only BN before the last layer separates the SSAD model $\mathcal{F}(\cdot; \bm{\theta})$ into two independent sub-models: $\mathcal{E}(\cdot; \bm{\theta}_\mcal{E})$ and $\mathcal{C}(\cdot; \bm{\theta}_\mcal{C})$, where $\mathcal{E}(\cdot; \bm{\theta}_\mcal{E})$ takes $X_\mcal{B}$ as inputs and outputs $H_\mcal{B} = \{\bm{h}_i\}_{i=1}^b$. $H_\mcal{B}$ is further normalized by BN, yielding $H'_\mcal{B} = \{\bm{h'}_i\}_{i=1}^b$.

\ssubsec{BN performs statistics towards normal data.}
Even though the corresponding loss supervision is removed, the normal data is still fed forward into the model $\mathcal{F}(\cdot; \bm{\theta})$ in our AnoOnly framework, as illustrated in Fig. \ref{fig:framework-anoonly}.
During forward propagation on mini-batch $\mcal{B}$, our AnoOnly utilizes BN to normalize the hidden feature $\bm{h}_i$ yielded from $\mathcal{E}(\bm{x}_i; \bm{\theta})$ and outputs
\begin{equation} \label{eq:bn}
  \bm{h'}_i = \frac{\bm{h}_i - \bm{\mu}_\mcal{B}}{\bm{\sigma}_\mcal{B}},
\end{equation}
where $\bm{\mu}_\mcal{B}$ and $\bm{\sigma}_\mcal{B}$ respectively indicate the mean and the standard deviation of $H_\mcal{B}$.
The $\bm{\mu}_\mcal{B}$ and $\bm{\sigma}_\mcal{B}$ quantitatively reflect the distribution of $H_\mcal{B}$, viewed as statistical characteristics of $\mathcal{B}$.
Since mini-batch $\mathcal{B}$ consistently adheres to the heavily imbalanced distribution in the entire dataset $\mathcal{D}$, the overwhelming majority or even the entirety of $\mathcal{B}$ comprises normal instances.
Hence, the statistical characteristics ($\bm{\mu}_\mcal{B}$ and $\bm{\sigma}_\mcal{B}$) gathered by BN primarily reflect ample normal data, with minimal impact from the few outliers (anomalies).

\ssubsec{BN performs online clustering learning.}
To intuitively disclose how BN introduces weak supervision with a form of cluster learning into our AnoOnly, we try to formulate the effect of BN as an explicit loss function.
Since $\bm{H'}_\mcal{B} = \{\bm{h'}_i\}_{i=1}^b$ is normalized by BN to approach the normal distribution $\mathcal{N}(\bm{0}, \bm{1})$, the mechanism of BN is empirically embodied as the following loss function:
\begin{align}
  L^\trm{BN}(\mathcal{D};\bm{\theta}_\mcal{E}) \
   & =  \sum_{\mcal{B}\subset\mcal{D}} \ \left(
  \left[ \frac{1}{b} \sum_{\mcal{B}} \bm{h'}_i - \bm{0} \right] + \
  \left[ \frac{1}{b-1} \sum_{\mcal{B}} (\bm{h'}_i-\bm{0})^2 - \bm{1} \right] \right) \nonumber
  \\
   & = \sum_{\mcal{B}\subset\mcal{D}} \ \left(
  \frac{1}{b} \sum_{\mcal{B}} \left[ \bm{h'}_i - \bm{0} \right] + \
  \frac{1}{b-1} \sum_{\mcal{B}} \left[ {\bm{h'}_i}^2 - \bm{1} \right] \right).
\end{align}
In order to reveal $L^\trm{BN}$ with respect to $\bm{h}$, we re-write it by introducing Eq. \ref{eq:bn} as follows:
\begin{align} \label{eq:L_BN}
  L^\trm{BN} & (\mathcal{D};\bm{\theta}_\mcal{E}) \nonumber \\
             & = \sum_{\mcal{B}\subset\mcal{D}} \ \left(
  \frac{1}{b} \sum_{\mcal{B}} \left[ \frac{\bm{h}_i - \bm{\mu}_\mcal{B}}{\bm{\sigma}_\mcal{B}} - \bm{0} \right] + \
  \frac{1}{b-1} \sum_{\mcal{B}} \left[ (\frac{\bm{h}_i - \bm{\mu}_\mcal{B}}{\bm{\sigma}_\mcal{B}})^2 - \bm{1} \right] \right)
  \\ \nonumber
             & = \sum_{\mcal{B}\subset\mcal{D}} \ \left(
  \frac{1}{b \, \bm{\sigma}_\mcal{B}} \sum_{\mcal{B}} \left[ \bm{h}_i - \bm{\mu}_\mcal{B} \right]
  + \frac{1}{(b-1)\bm{\sigma}_\mcal{B}^2} \sum_{\mcal{B}} \left[ (\bm{h}_i - \bm{\mu}_\mcal{B})^2 - (b-1)\bm{\sigma}^2_\mcal{B} \right] \right),
\end{align}
which is a standard loss form of clustering learning, and the clusters are a hypersphere with the center of $\bm{\mu}_\mcal{B}$ and radius of $\sqrt{(b-1)\bm{\sigma}_\mcal{B}^2}$.
The rationality of explicit $L^\trm{BN}$ is empirically justified by the ablation experiments in Section \ref{abla:bn}.
In particular, the hyperspherical clusters are controlled by $\bm{\mu}_\mcal{B}$ and $\bm{\sigma}_\mcal{B}$, which naturally updates throughout the model training.
Consequently, the cluster learning embedded in BN is an online process, in contrast to DeepSAD's superficial implementation with the fixed cluster center.
Correspondingly, by introducing $L^\trm{BN}$, the overall loss function in our AnoOnly can be explicitly presented as follows:
\begin{equation}
  L^\trm{AnoOnly}(\mathcal{D};\theta)=L^\trm{BN}(\mathcal{D};\bm{\theta}_\mcal{E})+L_A(\mathcal{D}_L;\bm{\theta})+\lambda R(\bm{\theta}),
\end{equation}
where $L^\trm{BN}$ provides weak supervision for the training of $\mathcal{E}(\cdot; \bm{\theta}_\mcal{E})$ by performing online cluster learning.

\ssubsec{BN attaches weak supervision alongside the strict supervision of loss $L_A$.}
Moreover, the statistical characteristics clustered by BN are inherently attached to the gradient during backward propagation as illustrated in Fig. \ref{fig:bn}.
Since we only resort to the implicit clustering of BN, the learnable scaling and shifting parameters in standard implementation \cite{ioffe2015batch} are overlooked, whose effect is empirically negligible as verified in Section \ref{abla:bn}.
According to the chain rule, with $\frac{\partial L_A}{\partial \bm{H'}_\mcal{B}}$ backward propagated, the gradient $\frac{\partial L_A}{\partial \bm{H}_\mcal{B}}$ yielded from BN is formulated as follows:
\begin{equation}
  \frac{\partial L_A}{\partial \bm{H}_\mcal{B}} = \
  \frac{\partial L_A}{\partial \bm{H'}_\mcal{B}} \cdot \frac{1}{\sqrt{\bm{\sigma}_\mcal{B}^2 + \bm{\epsilon}}} +
  \frac{\partial L_A}{\partial \bm{\sigma}_\mcal{B}^2} \cdot \frac{2(\bm{H}_\mcal{B}-\bm{\mu}_\mcal{B})}{b} +
  \frac{\partial L_A}{\partial \bm{\mu}_\mcal{B}} \cdot \frac{1}{b} ,
\end{equation}
where $\bm{\epsilon}$ is a small constant avoiding division by zero and the formulas of $\frac{\partial L_A}{\partial \bm{\sigma}_\mcal{B}^2}$ and $\frac{\partial L_A}{\partial \bm{\mu}_\mcal{B}}$ are provided in supplementary materials.
The explicit formulation of the gradient $\frac{\partial L_A}{\partial \bm{H}_\mcal{B}}$ intuitively demonstrates how BN incorporates the statistical characteristics primarily driven by normal data into the backpropagated gradient derived from $L_A$ on limited abnormal instances.

\ssubsec{BN performs as a divider to separate the training of $\mathcal{E}(\cdot; \bm{\theta}_\mcal{E})$ and $\mathcal{C}(\cdot; \bm{\theta}_\mcal{C})$.}
In our AnoOnly, BN not only structurally separates the model architecture of $\mathcal{E}(\cdot; \bm{\theta}_\mcal{E})$ and $\mathcal{C}(\cdot; \bm{\theta}_\mcal{C})$, but also assigns them distinct roles: feature enhancer ($\mathcal{E}$) and anomaly classifier ($\mathcal{C}$), respectively.
Due to the only explicit loss function $L_A$ in our $L^\trm{AnoOnly}$, the anomaly classifier $\mathcal{C}(\cdot; \bm{\theta}_\mcal{C})$ is trained to be only activated for abnormal data.
The loss supervision is incorporated with weak supervision of normal data when backward propagated through BN, enabling the representation learning of feature enhancer $\mathcal{E}(\cdot; \bm{\theta}_\mcal{E})$ for both normal and abnormal data.
As a result, our AnoOnly effectively leverages ample normal data to facilitate representation learning and shifts the model bias towards anomaly perception.
\section{Experiments}
We conduct comprehensive experiments across diverse datasets to validate the efficacy of our AnoOnly.
Empirical ablation studies are performed to validate and analyze the design of AnoOnly.
Additionally, we evaluate our AnoOnly on the contaminated dataset to assess the robustness of label noise.

\subsection{Datasets}
Our experiments are conducted on ten datasets, varying in data mode (image and text datasets), scale, and application domain.
All datasets are constructed with anomaly ratio $\frac{|\mathcal{D}_A|}{|\mathcal{D}|}=5\%$.
Following previous work \cite{han2022adbench}, all datasets are split into 70\% for training and 30\% for evaluation with the same normal/anomaly distribution.
The number of accessible anomaly data $\mathcal{D}_{LA}$ during training is controlled by the labeled anomaly ratio $\gamma_{la} = \frac{|\mathcal{D}_{LA}|}{|\mathcal{D}_A|}$.

\tbf{Image-based Datasets.} We evaluate our AnoOnly and prior works on five CV (Computer Vision) datasets. Among them, three classification datasets \emph{CIFAR10} \cite{krizhevsky2009cifar}, \emph{SVHN} \cite{netzer2011svhn}, and \emph{FashionMNIST} \cite{xiao2017fashion} are reconstructed by selecting one class as normal and other classes as abnormal.
As for \emph{MNIST-C} \cite{mu2019mnist}, standard MNIST data is set as normal while corrupted images are abnormal.
In \emph{MVTec-AD} \cite{bergmann2019mvtec} designed for industrial defect detection, $15$ types of industrial products are collected with accepts as normal and defects as abnormal.

\tbf{Text-based Datasets.} The other five NLP (Natural Language Processing) datasets are selected to highlight the generalization of our AnoOnly.
In sentiment analysis datasets \emph{Amazon} \cite{he2016amazon} and \emph{Imdb} \cite{maas2011imdb}, the negative comments are regarded as the abnormal class.
For \emph{Yelp} \cite{he2016amazon}, the reviews of 0 and 1 stars are set as the abnormal class and the reviews with more stars are normal data.
As for the topic classification datasets \emph{20newsgroups} \cite{lang1995newsgroup} and \emph{Agnews} \cite{zhang2015agnews}, we also set one class as normal and downsample instances of the remaining classes as anomalies.

\subsection{Evaluation protocol}
We calculate the widely-used AUCROC (Area Under Receiver Operating Characteristic Curve) and AUCPR (Area Under Precision-Recall Curve) to evaluate the detection performance following previous works \cite{han2022adbench,akcay2019ganomaly,ruff2019deep,bergmann2019mvtec}. In particular, the AUCPR score with respect to (w.r.t.) anomaly visually evaluates the detection accuracy targeting anomalies.

\subsection{Implementation details}
Our AnoOnly framework seamlessly integrates into existing SSAD methods, with no additional burden. 
Besides DeepSAD \cite{ruff2019deep}, we validate AnoOnly on the other $3$ SSAD methods: DevNet \cite{pang2019devnet}, PReNet \cite{pang2019deep} and FEAWAD\cite{zhou2021feature} to assess its model generalization.
To achieve the integration, we discontinue the loss supervision on normal in these methods and introduce a batch normalization layer before the final output layer if no original BN is present, without other adjustments. Comprehensive algorithmic details and hyperparameter settings can be found in the supplementary materials.
For a fair comparison, we employ the same pre-trained backbone (ResNet-18 \cite{he2016resnet} for CV datasets and Bert \cite{devlin2018bert} for NLP datasets) for feature extraction. Subsequently, we exclusively train SSAD models (anomaly detection heads), using identical extracted features, enabling an objective evaluation of different SSAD methods. Given that training anomaly detection heads composed of a few fully connected layers is sufficient, our experiments can be conveniently executed on commercially available GPUs or even CPUs, thereby achieving computational feasibility.

\begin{table}[t]
  \caption{Overall performance (AUCROC in $\%$) comparison on ten anomaly detection datasets with labeled anomaly ratio $\gamma_{la}=10\%$. \textcolor{blue}{$_{\uparrow\left(\cdot\right)}$} indicates performance enhancement with AnoOnly integrated.}
  \label{tab:comparison}
  \centering
  \tabcolsep=7pt
  \resizebox{\linewidth}{!}{%
    \begin{tabular}{@{}lp{1.1cm}<{\centering}p{1.1cm}<{\centering}p{1.1cm}<{\centering}p{1.1cm}<{\centering}p{1.1cm}<{\centering}p{1.1cm}<{\centering}p{1.1cm}<{\centering}p{1.1cm}<{\centering}p{1.1cm}<{\centering}p{1.1cm}<{\centering}p{1.1cm}<{\centering}@{}}
      \toprule
      \multicolumn{1}{c}{\multirow{2}{*}[-3pt]{Methods}} & \multicolumn{5}{c}{Image-based (CV) datasets} & \multicolumn{5}{c}{Text-based (NLP) datasets} & \multirow{2}{*}[-3pt]{\begin{tabular}[c]{@{}c@{}}Overall\\ Avg.\end{tabular}}                                                                                                                                                                     \\ \cmidrule(lr){2-6} \cmidrule(lr){7-11}
      \multicolumn{1}{c}{}                               & Cifar10                                       & FMnist                                        & MnistC                                            & MVTec             & SVHN              & 20news            & Agnews            & Amazon            & Imdb              &  Yelp               &                   \\ \midrule
      {GANomaly\cite{akcay2019ganomaly}} & 66.1  & 78.2  & 75.0  & 75.8  & 56.5  & 54.3  & 53.7  & 56.8  & 48.8  & 54.6  & 62.0  \\
    {REPEN\cite{pang2018learning}} & 66.8  & 87.0  & 80.0  & 74.2  & 58.8  & 56.6  & 60.6  & 58.2  & 57.1  & 63.9  & 66.3  \\
    {XGBOD*\cite{zhao2018xgbod}} & 78.9  & 92.7  & 92.8  & 80.8  & 72.8  & 69.0  & 83.5  & 73.2  & 77.6  & 80.0  & 80.1  \\ \midrule
    {DeepSAD\cite{ruff2019deep}} & 64.4  & 83.6  & 74.0  & 69.3  & 60.1  & 49.2  & 64.9  & 68.1  & 51.4  & 68.5  & 65.4  \\
    {+AnoOnly} & \tbf{82.7}\diffH{18.3} & \tbf{94.1}\diffH{10.5} & \tbf{94.3}\diffH{20.3} & \tbf{77.9}\diffH{8.6} & \tbf{76.4}\diffH{16.4} & \tbf{69.9}\diffH{20.7} & \tbf{91.2}\diffH{26.3} & \tbf{85.9}\diffH{17.7} & \tbf{85.6}\diffH{34.2} & \tbf{91.7}\diffH{23.2} & \tbf{85.0}\diffH{19.6} \\ \midrule
    {FEAWAD\cite{zhou2021feature}} & 70.6 & 87.9 & 88.4 & 63.3 & 66.5 & 67.9 & 81.5 & 77.8 & 79.6 & 84.2 & 76.8 \\
    {+AnoOnly} & \tbf{79.8}\diffH{9.2} & \tbf{93.6}\diffH{5.8} & \tbf{93.5}\diffH{5.1} & \tbf{75.5}\diffH{12.2} & \tbf{72.2}\diffH{5.6} & \tbf{75.6}\diffH{7.7} & \tbf{88.6}\diffH{7.1} & \tbf{80.0}\diffH{2.2} & \tbf{82.7}\diffH{3.1} & \tbf{86.5}\diffH{2.3} & \tbf{82.8}\diffH{6.0} \\ \midrule
    {PReNet\cite{pang2019deep}} & 75.1  & 92.3  & 94.0  & 74.4  & \tbf{73.0}  & 75.1  & 87.4  & 83.5  & 81.4  & 88.3  & 82.4  \\
    {+AnoOnly} & \tbf{76.9}\diffH{1.8}  & \tbf{93.0}\diffH{0.7}  & \tbf{94.4}\diffH{0.4}  & \tbf{78.8}\diffH{4.3}  & 72.8\diffL{-0.2}  & \tbf{78.6}\diffH{3.5}  & \tbf{91.8}\diffH{4.4}  & \tbf{85.1}\diffH{1.7}  & \tbf{86.2}\diffH{4.8}  & \tbf{89.7}\diffH{1.4}  & \tbf{84.7}\diffH{2.3}  \\ \midrule
    {DevNet\cite{pang2019devnet}} & \tbf{81.3}  & 92.8  & 93.9  & \tbf{77.6}  & 74.4  & 68.3  & 91.3  & 85.8  & \tbf{86.8}  & 91.5  & 84.4  \\
    {+AnoOnly} & 80.4\diffL{-0.9}  & \tbf{93.7}\diffH{0.9}  & \tbf{94.6}\diffH{0.7}  & 76.8\diffL{-0.7}  & \tbf{76.1}\diffH{1.7}  & \tbf{74.5}\diffH{6.2}  & \tbf{92.0}\diffH{0.7}  & \tbf{86.2}\diffH{0.4}  & 85.6\diffL{-1.2}  & \tbf{91.6}\diffH{0.0}  & \tbf{85.2}\diffH{0.8}\\
      \bottomrule
    \end{tabular}%
  }
\end{table}

\subsection{Performance enhancements when integrating into state-of-the-arts}
\begin{figure*}[h]
  \begin{minipage}[b][][b]{0.48\linewidth}
    \centering
    \includegraphics[width=0.98\linewidth]{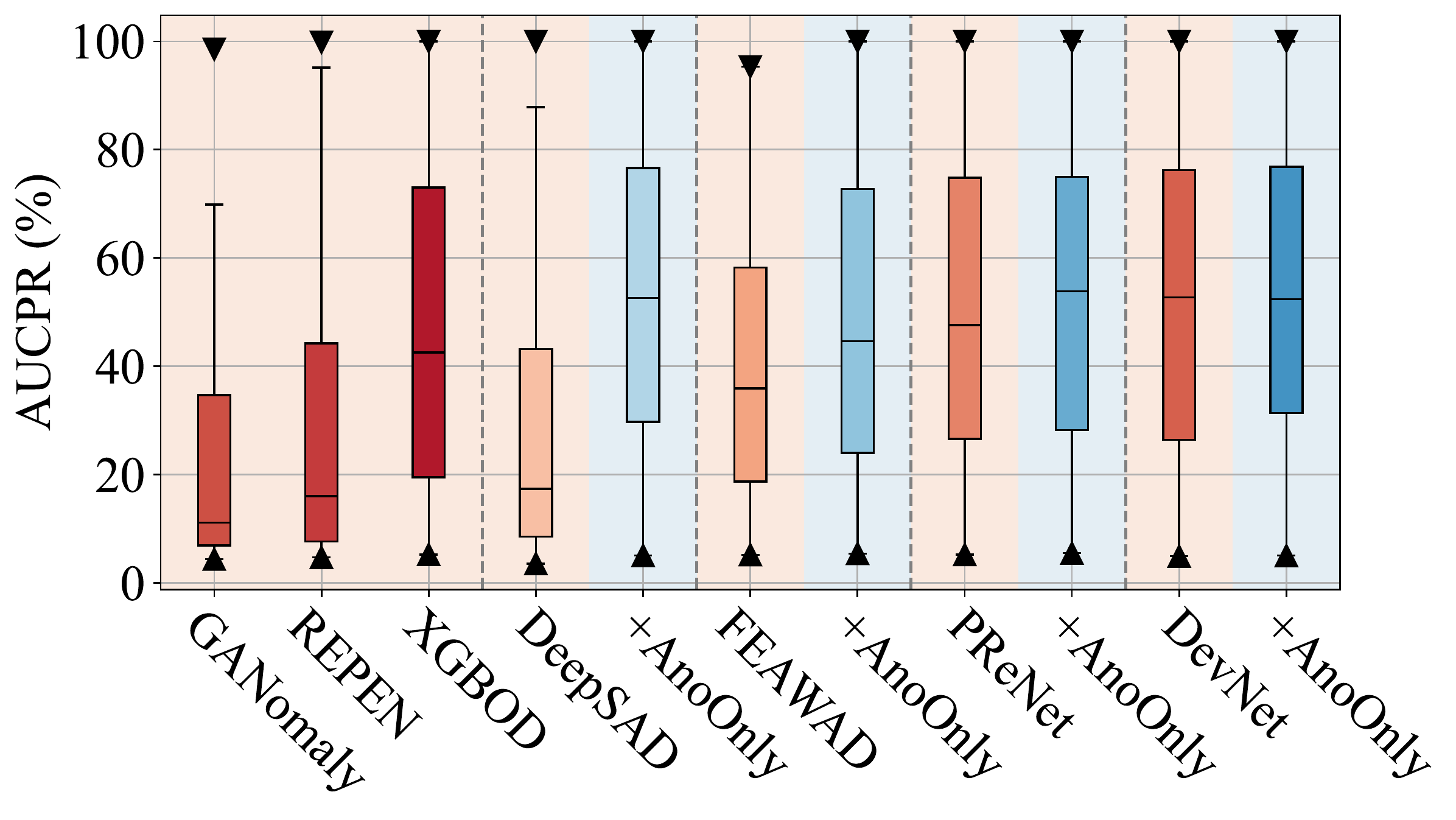}
    \caption{Boxplot of AUCPR ($\%$) scores on ten datasets with labeled anomaly ratio $\gamma_{la}$=$10\%$. We denote prior SSAD methods in $\textcolor{bg-ssad}{\blacksquare}$ and those incorporated with our AnoOnly in $\textcolor{bg-anoonly}{\blacksquare}$.}
    \label{fig:boxplot}
  \end{minipage}
  \hfill
  \begin{minipage}[b][][b]{0.49\linewidth}
    \includegraphics[width=0.99\linewidth]{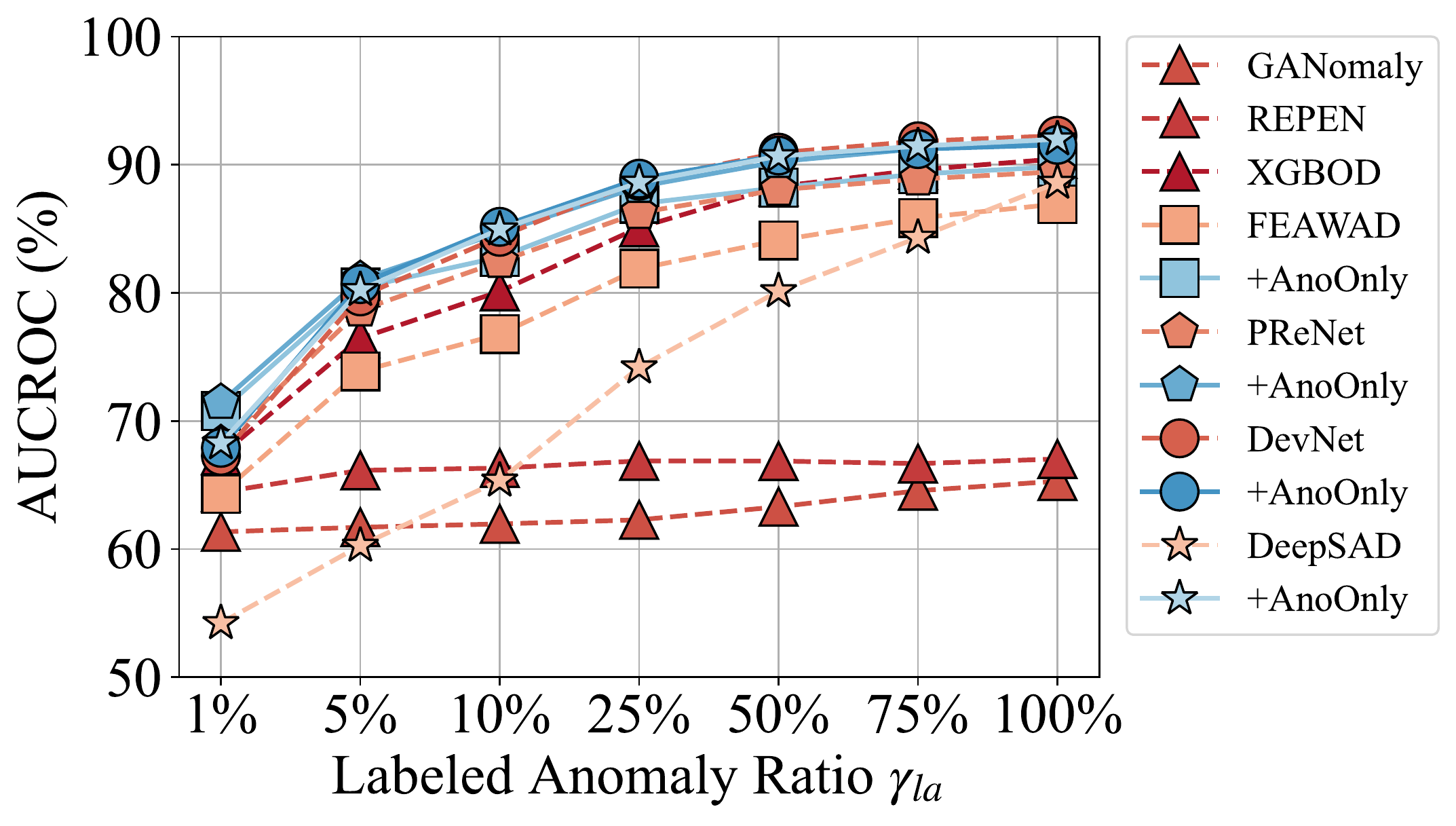}
    \caption{Overall AUCROC ($\%$) variation with increasing labeled anomaly ratio $\gamma_{la}$. Dashed lines are used for prior methods and solid lines indicate those incorporated with our AnoOnly.}
    \label{fig:line-plot-aucroc}
  \end{minipage}
\end{figure*}

We first present a performance comparison of prior SSAD methods with or without our AnoOnly, under the setting of labeled anomaly ratio $\gamma_{la}=10\%$ for all datasets.
As shown in Table \ref{tab:comparison} providing detailed dataset-wise AUCROC, the proposed AnoOnly improves the performance of all incorporated methods across almost all datasets of various domains, verifying its efficacy and excellent generalization capability.
Specifically, when combined with DeepSAD \cite{ruff2019deep}, our AnoOnly boosts it to outperform all existing SSAD methods, yielding an impressive overall AUCROC gain of up to $19.6\%$.
As for other $3$ methods \cite{pang2019devnet,zhou2021feature,pang2019deep}, they rebalance the one-sided supervision volume by over-sampling limited abnormal data to amplify the supervision volume targeting anomalies.
The proposed AnoOnly further elevates them to achieve new SOTA performance by decreasing the excessive supervision volume on normal data.
Furthermore, the boxplot of AUCPR as illustrated in Fig. \ref{fig:boxplot} provides a visual representation of consistent improvements benefited from our AnoOnly in detection accuracy (AUCPR) with respect to the anomaly.

To assess the efficacy of the proposed AnoOnly from a more comprehensive perspective, we report the overall performance variation with an increasing labeled anomaly ratio $\gamma_{la} \in [1\%, 5\%, 10\%, 25\%, 50\%, 75\%, 100\%]$ in Fig. \ref{fig:line-plot-aucroc}.
When only a handful of anomalies are labeled ($\gamma_{la} \in [1\%, 5\%, 10\%, 25\%]$) during training, our AnoOnly framework demonstrates remarkable superiority in taking full advantage of insufficient abnormal data.
Although the superiority of AnoOnly diminishes as more abnormal data being accessible for training, it is worth noting that collecting and annotating a significant amount of abnormal instances is impractical in real-world scenarios.

\begin{table}[htbp]
  \caption{AUCPR w.r.t. anomaly ($\%$) comparison on ten anomaly detection datasets with labeled anomaly ratio $\gamma_{la}=10\%$. \textcolor{blue}{$_{\uparrow\left(\cdot\right)}$} denotes $\trm{AUCPR}_\trm{anomaly}$ enhancement with AnoOnly integrated.}
  \label{tab:comparison_pr}
  \centering
  \tabcolsep=7pt
  \resizebox{\linewidth}{!}{
      \begin{tabular}{@{}lp{1.1cm}<{\centering}p{1.1cm}<{\centering}p{1.1cm}<{\centering}p{1.1cm}<{\centering}p{1.1cm}<{\centering}p{1.1cm}<{\centering}p{1.1cm}<{\centering}p{1.1cm}<{\centering}p{1.1cm}<{\centering}p{1.1cm}<{\centering}p{1.1cm}<{\centering}@{}}
          \toprule
          \multicolumn{1}{c}{\multirow{2}{*}[-3pt]{Methods}} & \multicolumn{5}{c}{Image-based (CV) datasets} & \multicolumn{5}{c}{Text-based (NLP) datasets} & \multirow{2}{*}[-3pt]{\begin{tabular}[c]{@{}c@{}}Overall\\ Avg.\end{tabular}}                                   \\ \cmidrule(lr){2-6} \cmidrule(lr){7-11}
          \multicolumn{1}{c}{}                         &
          Cifar10                                       & FMnist                                       & MnistC                                    & MVTec & SVHN & 20news & Agnews & Amazon & Imdb & \, Yelp &                                  \\ \midrule
          {GANomaly \cite{akcay2019ganomaly}}                                     & 11.1  & 26.1  & 23.2  & 58.0  & 7.2   & 7.1   & 6.3   & 6.4   & 5.1   & 6.3   & 15.7  \\     
          {REPEN \cite{pang2018learning}}                                        & 12.1  & 41.7  & 33.7  & 55.7  & 8.3   & 7.1   & 7.6   & 6.0   & 6.2   & 10.1  & 18.8  \\
          {XGBOD \cite{zhao2018xgbod}}                                       & 24.4  & 64.8  & 74.3  & 69.0  & 18.0  & 15.2  & 30.3  & 15.0  & 17.6  & 25.1  & 35.4  \\ \midrule
          {DeepSAD \cite{ruff2019deep}}                                      & 10.6  & 50.6  & 37.6  & 53.3  & 8.7   & 7.4   & 9.7   & 8.0   & 5.0   & 11.7  & 20.3  \\
          {+AnoOnly} & \,32.5\diffH{21.9} & \,75.9\diffH{25.3} & \,80.9\diffH{43.3} & \,64.5\diffH{11.2} & \,22.2\diffH{13.5} & \,20.7\diffH{13.3} & \,59.2\diffH{49.5} & \,33.7\diffH{25.7} & \,35.5\diffH{30.5} & \,50.4\diffH{38.7} & \,47.6\diffH{27.3} \\ \midrule
        {FEAWAD \cite{zhou2021feature}} & 18.9 & 62.9 & 59.0 & 50.8 & 13.9 & 15.6 & 35.6 & 18.2 & 19.2 & 29.6 & 32.4 \\
        {+AnoOnly} & \,27.0\diffH{8.0} & \,71.3\diffH{8.4} & \,77.1\diffH{18.1} & \,62.6\diffH{11.8} & \,18.1\diffH{4.1} & \,22.6\diffH{7.0} & \,44.9\diffH{9.3} & \,25.6\diffH{7.4} & \,24.4\diffH{5.1} & \,41.6\diffH{12.0} & \,41.5\diffH{9.1} \\ \midrule
        {PReNet \cite{pang2019deep}} & 26.9 & 74.5 & 80.6 & 61.1 & 20.0 & 22.4 & 38.7 & 29.1 & 20.0 & 34.9 & 40.8 \\
        {+AnoOnly} & \,26.0\diffL{-0.9} & \,73.4\diffL{-1.1} & \,82.2\diffH{1.6} & \,65.9\diffH{4.8} & \,19.5\diffL{-0.5} & \,27.0\diffH{4.6} & \,54.6\diffH{15.9} & \,33.0\diffH{3.9} & \,34.9\diffH{14.9} & \,45.6\diffH{10.7} & \,46.2\diffH{5.4} \\ \midrule
        {DevNet \cite{pang2019devnet}} & 27.2 & 74.5 & 81.4 & 63.7 & 21.2 & 15.0 & 57.8 & 35.1 & 33.6 & 49.2 & 45.9 \\
        {+AnoOnly} & \,29.3\diffH{2.1} & \,74.5\diffH{0.0} & \,82.1\diffH{0.7} & \,62.4\diffL{-1.4} & \,22.1\diffH{0.9} & \,26.1\diffH{11.1} & \,59.0\diffH{1.2} & \,39.2\diffH{4.1} & \,35.3\diffH{1.7} & \,52.0\diffH{2.8} & \,48.2\diffH{2.3}
                                                 \\ \bottomrule
      \end{tabular}
  }
\end{table}
\begin{table}[htbp]
  \caption{AUCPR w.r.t. normal ($\%$) comparison on ten anomaly detection datasets with labeled anomaly ratio $\gamma_{la}=10\%$. \textcolor{red}{$_{\downarrow\left(\cdot\right)}$} denotes $\trm{AUCPR}_\trm{normal}$ degradation with AnoOnly integrated.}
  \label{tab:comparison_prnormal}
  \centering
  \tabcolsep=7pt
  \resizebox{\linewidth}{!}{
      \begin{tabular}{@{}lp{1.1cm}<{\centering}p{1.1cm}<{\centering}p{1.1cm}<{\centering}p{1.1cm}<{\centering}p{1.1cm}<{\centering}p{1.1cm}<{\centering}p{1.1cm}<{\centering}p{1.1cm}<{\centering}p{1.1cm}<{\centering}p{1.1cm}<{\centering}p{1.1cm}<{\centering}@{}}
          \toprule
          \multicolumn{1}{c}{\multirow{2}{*}[-3pt]{Methods}} & \multicolumn{5}{c}{Image-based (CV) datasets} & \multicolumn{5}{c}{Text-based (NLP) datasets} & \multirow{2}{*}[-3pt]{\begin{tabular}[c]{@{}c@{}}Overall\\ Avg.\end{tabular}}                                   \\ \cmidrule(lr){2-6} \cmidrule(lr){7-11}
          \multicolumn{1}{c}{}                         & Cifar10                                       & FMnist                                       & MnistC                                    & MVTec & SVHN & 20news & Agnews & Amazon & Imdb & \, Yelp &      \\ \midrule
          {GANomaly \cite{akcay2019ganomaly}}          & 92.4  & 89.9  & 90.3  & 64.4  & 93.9  & 94.7  & 94.3  & 94.0  & 95.1  & 94.1  & 90.3  \\
          {REPEN \cite{pang2018learning}}              & 92.0  & 87.4  & 89.0  & 65.1  & 93.4  & 94.4  & 93.5  & 94.2  & 94.1  & 92.4  & 89.6  \\
          {XGBOD \cite{zhao2018xgbod}}                & 89.1  & 85.7  & 85.9  & 62.1  & 90.5  & 92.0  & 88.2  & 90.5  & 89.9  & 88.9  & 86.3  \\ \midrule
          {DeepSAD \cite{ruff2019deep}}                & 92.5  & 87.6  & 89.8  & 66.3  & 93.3  & 94.9  & 92.5  & 92.9  & 95.2  & 91.7  & 89.7  \\
          {+AnoOnly}                                   & \,88.2\diffL{-4.3}  & \,85.3\diffL{-2.2}  & \,85.5\diffL{-4.3}  & \,63.2\diffL{-3.1}  & \,89.8\diffL{-3.5}  & \,91.3\diffL{-3.6}  & \,86.1\diffL{-6.4}  & \,87.6\diffL{-5.3}  & \,87.5\diffL{-7.6}  & \,86.1\diffL{-5.5}  & \,85.1\diffL{-4.6}  \\ \midrule
          {FEAWAD \cite{zhou2021feature}}              & 90.6  & 86.3  & 86.7  & 67.6  & 91.6  & 91.6  & 88.0  & 89.7  & 89.5  & 88.0  & 87.0  \\
          {+AnoOnly}                                   & \,88.9\diffL{-1.7}  & \,85.5\diffL{-0.9}  & \,85.7\diffL{-1.0}  & \,63.5\diffL{-4.1}  & \,90.8\diffL{-0.8}  & \,90.8\diffL{-0.8}  & \,86.9\diffL{-1.2}  & \,88.8\diffL{-0.9}  & \,88.6\diffL{-0.9}  & \,87.2\diffL{-0.8}  & \,85.6\diffL{-1.3}  \\ \midrule
          {PReNet \cite{pang2019deep}}                 & 89.2  & 85.5  & 85.5  & 63.8  & 90.4  & 90.7  & 87.3  & 88.1  & 89.2  & 87.3  & 85.7  \\
          {+AnoOnly}                                   & \,89.2\diffH{0.0}  & \,85.5\diffH{0.0}  & \,85.5\diffL{-0.1}  & \,62.4\diffL{-1.4}  & \,90.4\diffH{0.1}  & \,89.8\diffL{-0.9}  & \,86.1\diffL{-1.1}  & \,87.7\diffL{-0.5}  & \,87.5\diffL{-1.7}  & \,86.5\diffL{-0.7}  & \,85.1\diffL{-0.6}  \\ \midrule
          {DevNet \cite{pang2019devnet}}               & 88.8  & 85.6  & 85.6  & 63.1  & 90.2  & 92.2  & 86.1  & 87.6  & 87.5  & 86.2  & 85.3  \\
          {+AnoOnly}                                   & \,88.6\diffL{-0.1}  & \,85.4\diffL{-0.2}  & \,85.4\diffL{-0.1}  & \,63.2\diffH{0.1}  & \,89.8\diffL{-0.4}  & \,90.1\diffL{-2.1}  & \,86.0\diffL{-0.2}  & \,87.3\diffL{-0.3}  & \,87.6\diffH{0.1}  & \,86.1\diffL{-0.1}  & \,85.0\diffL{-0.3}  \\ \bottomrule

      \end{tabular}
  }
\end{table}

\subsection{AUCPR towards anomaly and normal}
We report the performance comparison of AUCPR for anomaly ($\trm{AUCPR}_\trm{anomaly}$) and normal ($\trm{AUCPR}_\trm{normal}$) on ten diverse datasets in Table \ref{tab:comparison_pr} and Table \ref{tab:comparison_prnormal}, respectively.
From the results in Table \ref{tab:comparison_pr}, the proposed AnoOnly effectively facilitates the detection precision towards anomalies when incorporating with existing methods \cite{pang2019deep,pang2019devnet,ruff2019deep,zhou2021feature}. In particular, our AnoOnly achieves an overall $\trm{AUCPR}_\trm{anomaly}$ improvement of up to $27.3\%$ for the vanilla DeepSAD \cite{ruff2019deep} without rebalancing strategy. As for the other three methods resorting to over-sampling for abnormal instances, notable performance enhancements 
are also observed on most datasets when incorporating the AnoOnly framework. These significant improvements in $\trm{AUCPR}_\trm{anomaly}$ convincingly demonstrate the successful redirection of trained models towards anomaly perception achieved by AnoOnly.

Faultily, as shown in Table \ref{tab:comparison_prnormal}, due to the absence of explicit loss supervision on normal data, our AnoOnly incurs a decrease in detection precision for normal instances.
Nevertheless, considering the substantial improvements in anomaly detection, the relatively modest performance degradation in normal detection is acceptable. Additionally, the AUCROC comparison presented in the main text indicates positive improvements in the trade-off between detection performance for normal and abnormal instances. Moreover, in many practical anomaly detection applications such as cancer diagnosis, the cost of missing cancer tumors (anomalies) is much higher than misclassifying benign tumors (normal) as cancer tumors.

\subsection{Empirical ablation studies}

In the ablation studies, we first evaluate the effectiveness of the weak supervision mechanism introduced by BN for normal data in our AnoOnly framework.
Subsequently, we elaborately design ablation experiments to verify that BN's property of capturing statistical characteristics and implicitly introducing online cluster learning plays a crucial role in the efficacy of AnoOnly.
For all ablation experiments, we employ the vanilla DeepSAD without re-sampling strategy as the baseline to integrate into our AnoOnly.
The AUCROC and AUCPR scores averaged on 10 datasets are reported, where the labeled anomaly ratio $\gamma_{la}$ is set at $10\% $ to meet the few-shot setting of SSAD.

\begin{figure*}[t]
  \begin{minipage}[b][][b]{0.48\linewidth}
    \centering
    \includegraphics[width=\linewidth]{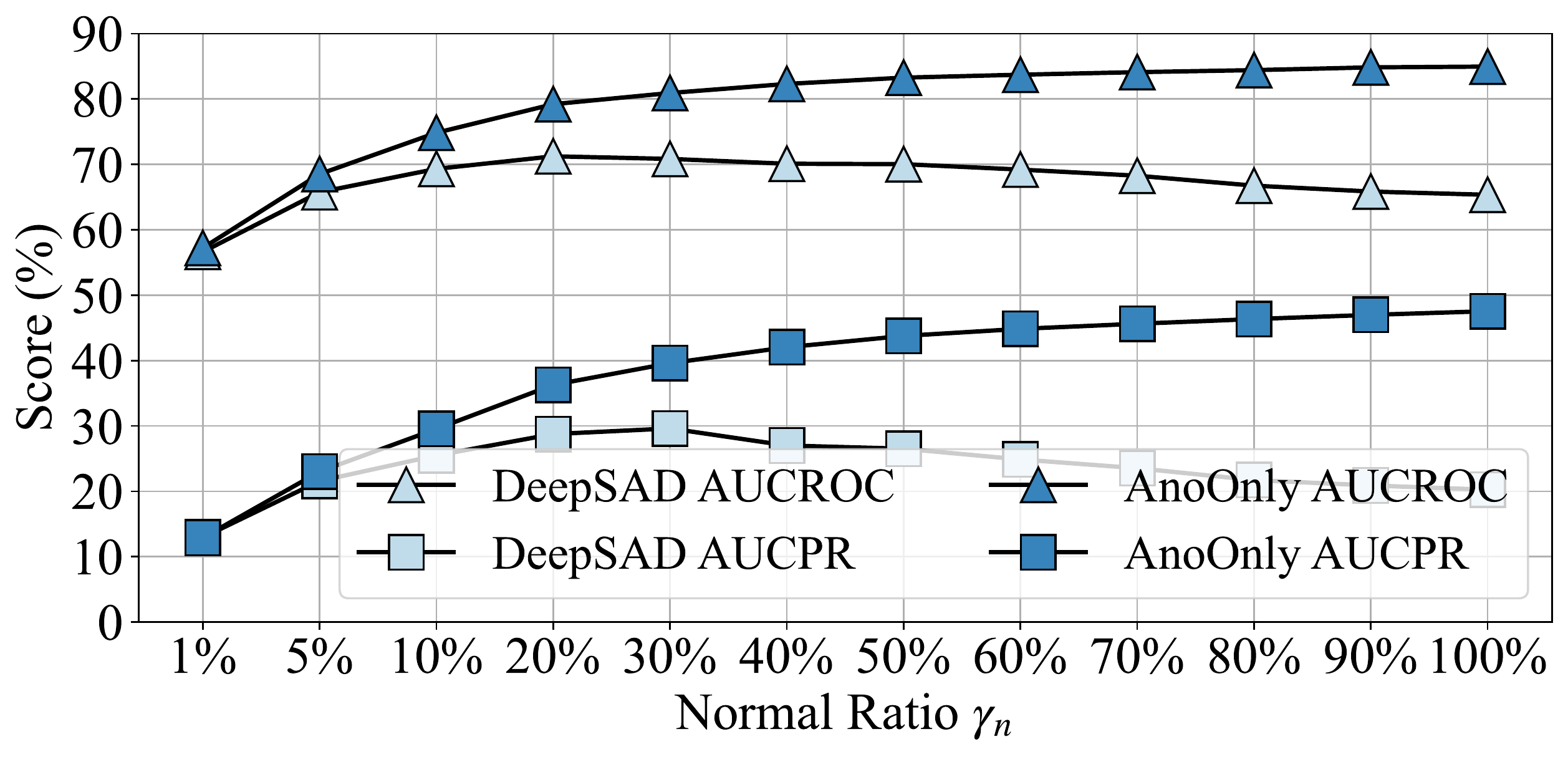}
    \caption{AUCROC $(\%)$ and AUCPR $(\%)$ variation of DeepSAD \cite{ruff2019deep} and AnoOnly with increasing normal ratio $\gamma_n$.}
    \label{fig:abla-normal-num}
  \end{minipage}
  \hfill
  \begin{minipage}[b][][b]{0.48\linewidth}
    \centering
    \includegraphics[width=\linewidth]{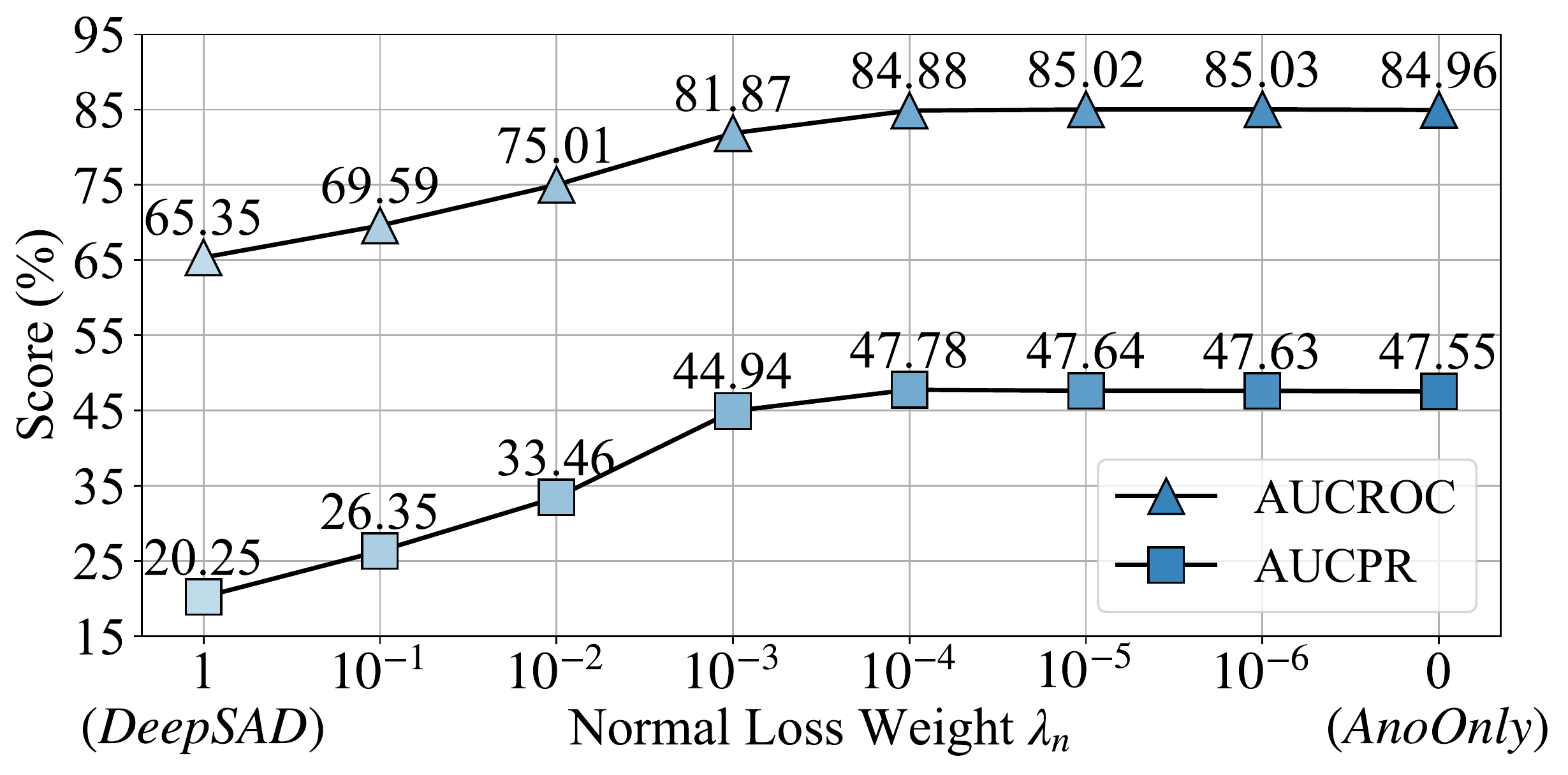}
    \caption{AUCROC $(\%)$ and AUCPR $(\%)$ variation with decreasing normal loss weight $\lambda_n$ from $1$ (\emph{DeepSAD} \cite{ruff2019deep}) to $0$ (\emph{AnoOnly}).}
    \label{fig:abla-normal-weight} 
  \end{minipage}
\end{figure*}

\ssubsec{The effect of the varying data amount of normal data.}
Although the explicit loss supervision on normal data is excluded in our AnoOnly, its forward propagation remains crucial for implicit cluster learning through the utilization of BN, as emphasized in Section \ref{sec:bn}.
To highlight the superiority of weak supervision mechanism towards normal data in AnoOnly, we conduct ablation on DeepSAD and AnoOnly with increasing normal ratio $\gamma_n$ as shown in Fig. \ref{fig:abla-normal-num}.
When $\gamma_n$ is reduced to less than $10\%$, both DeepSAD and AnoOnly suffer from intolerable performance deterioration, demonstrating the indispensability of normal data for existing SSAD methods and our AnoOnly.

Particularly, the performance of DeepSAD instead gradually declines with $\gamma_n$ growing from $20\%$ to $100\%$, especially in AUCPR towards detection of anomalies.
This performance degradation verifies that the overwhelming supervision volume on normal data suppressing anomalies significantly biases model attention and capacity against anomaly perception.
As for our AnoOnly leveraging BN to introduce weak supervision on normal datasets to rebalance supervision volume, the increasing abundance of normal data brings continuous performance improvements, where BN captures more comprehensive statistical characteristics.

\ssubsec{The effect of the varying loss weight of $L_N$ for normal data.}
The relative loss supervision volume between normal and abnormal data can also be controlled by introducing a hyperparameter $\lambda_n$ to re-weight $L_N$ for normal data.
From this perspective, DeepSAD and our AnoOnly are interpreted as two extreme instances of the re-weighting scheme, whose $\lambda_n$ is assigned as $1$ or reduced to $0$.
Varying $\lambda_n$ builds compromises between DeepSAD and AnoOnly, enabling a closer look at the supervision volume balance.
As illustrated in Fig. \ref{fig:abla-normal-weight}, when we reduce $\lambda_n=1$ of the standard DeepSAD to $1e-6$, significant performance improvements are observed. 
These performance gains further substantiate our insight that the superfluous supervision assigned to normal data in existing SSAD methods \cite{zhao2018xgbod,pang2019devnet,zhou2021feature,pang2019deep} dramatically biases the model's attention.

In particular, retaining minimal loss supervision on normal data ($\lambda_n=10^{-5}$, $10^{-6}$) demonstrates slight improvements over our extreme AnoOnly in both AUCROC and AUCPR.
We present this observation as an inspiring insight for future research, suggesting the exploration of alternative approaches to implement controllable weak supervision on normal data to further facilitate supervision rebalancing between heavily imbalanced normal and abnormal data.

\ssubsec{The ablation studies on batch normalization.} \label{abla:bn}
In this ablation study, we delve into the efficacy of batch normalization for our AnoOnly framework by replacing BN with tailored variants as demonstrated in Table \ref{tab:bn-variants}.

\begin{wrapfigure}{R}{0.5\linewidth}
  \centering
  \vspace{-13pt}
  \captionof{table}{Ablation studies on BN. `w/o BN' denotes removing BN in models. `LN' denotes layer normalization \cite{ba2016layer}. `BN*' denotes BN without affine parameters. `BN$^\dagger$' denotes the manual implementation of $L^{BN}$. \textcolor{blue}{$_{\uparrow\left(\cdot\right)}$} denotes AUCROC ($\%$) gains over `w/o BN'.}
  \label{tab:bn-variants}
  \resizebox{0.4\textwidth}{!}{%
  \begin{tabular}{@{}ccc@{}}
    \toprule
    AUCROC ($\%$)                    & DeepSAD \cite{ruff2019deep} & +AnoOnly              \\ \midrule
    w/o BN                    & 61.16 \,\,                       & 53.02                  \\
    LN \cite{ba2016layer}     & 65.29 \diffH{4.13}          & \,\,\,53.85   \diffH{0.83}  \\
    BN* \cite{ioffe2015batch} & 65.35 \diffH{0.79}          & \,\,\,84.96   \diffH{31.91} \\
    BN  \cite{ioffe2015batch} & 61.95 \diffH{4.19}          & \,\,\,84.93   \diffH{31.94} \\
    BN$^\dagger$              & -                           & \,\,\,84.61   \diffH{31.59} \\
    \bottomrule
  \end{tabular}%
}
  \vspace{-5pt}
\end{wrapfigure}

\emph{1) BN v.s. w/o BN.} When the only batch normalization layer is removed from the model (w/o BN), the overall AUCROC score of DeepSAD experiences a slight reduction of  $4.19\%$. However, as for our AnoOnly without BN, the trained model suffers from a collapse, resulting in a drastic performance degradation of up to $31.94\%$.
This stark contrast underscores the indispensability of BN in the design of our AnoOnly framework.

\emph{2) BN v.s. LN.} In order to eliminate the stabilizing effect inherent in BN, we replace it with layer normalization (LN), which assumes the same role in improving training stabilization.
However, since LN performs statistics along the other dimension, it is unable to capture the comprehensive statistical characteristics of abundant normal data, unlike BN.
Although the additional LN brings marginal performance gains, there remains a substantial performance decline compared to the presence of BN in AnoOnly.
This remaining performance gap verifies that the effect of BN in stabilizing training is not the crucial factor for the superiority of our AnoOnly.

\emph{3) BN v.s. BN*.} We disable the learnable affine parameters in standard BN and denote it as `BN*'. As for the vanilla DeepSAD, this removal leads to an obvious performance drop ($3.40\%$). However, this modification demonstrates negligible effect for our AnoOnly.

\emph{4) BN v.s. BN$^\dagger$.} Through a gradual dissection of BN, we empirically identify its pivotal role in our AnoOnly, which is capturing statistical characteristics of normal data and performing online cluster learning, as elaborated in Section \ref{sec:bn}.
To further justify the rationality of explicit cluster learning loss $L_{BN}$ (Eq. \ref{eq:L_BN}) derived from BN, we disable BN in AnoOnly and manually incorporate this loss function, denoted as BN$^\dagger$.
This replacement demonstrates a negligible impact on overall performance ($84.93\%$ v.s. $84.61\%$), thereby affirming the essential function of BN in the AnoOnly framework.

\ssubsec{The effect of batch size for our AnoOnly}

\begin{wrapfigure}{R}{0.5\linewidth}
  \centering
  \includegraphics[width=1.0\linewidth]{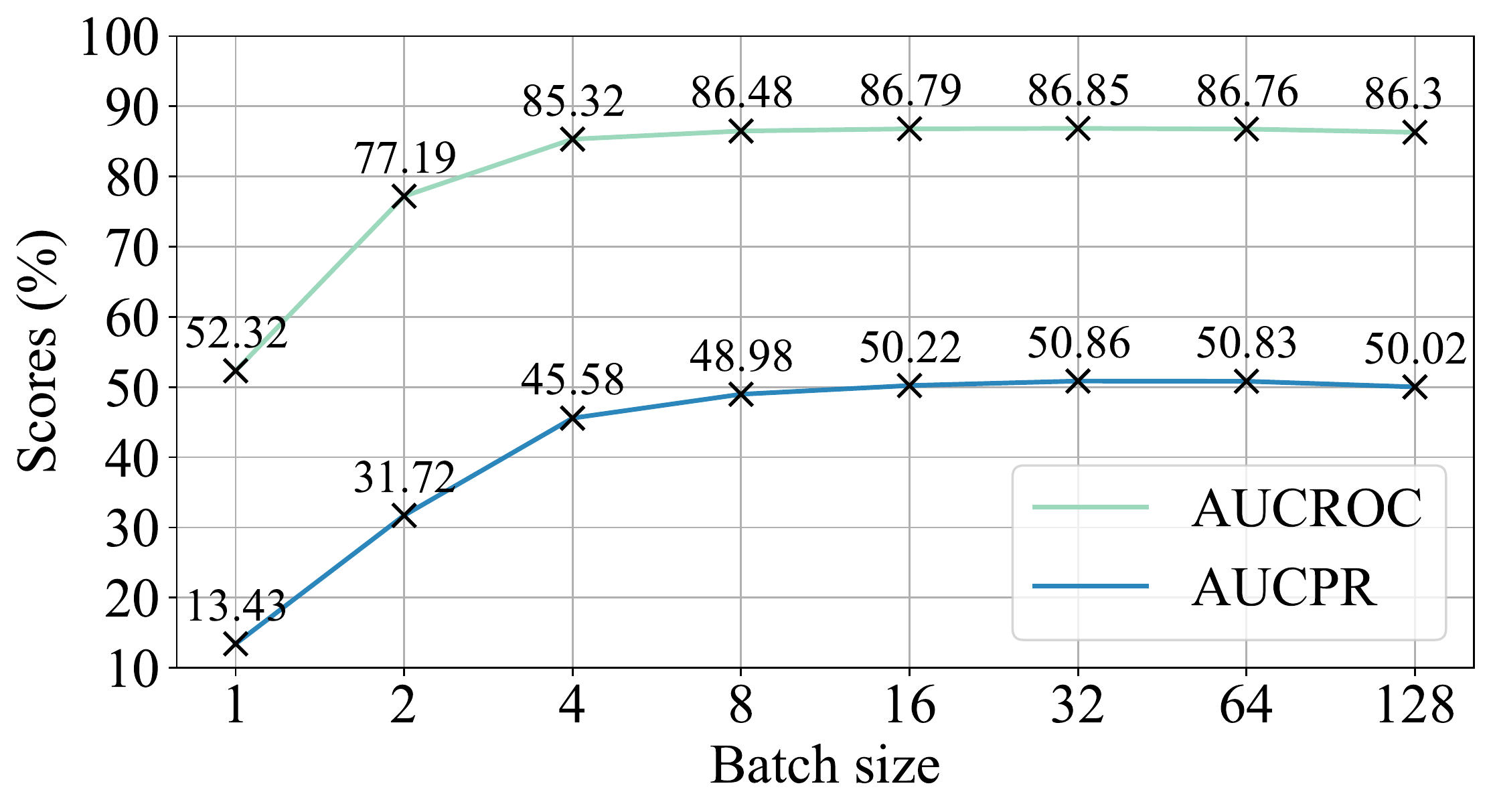}
  \caption{Overall AUCROC ($\%$) and AUCPR ($\%$) variation of DeepSAD \cite{ruff2019deep} +AnoOnly with increasing batch size. }
  \label{fig:bz}
\end{wrapfigure}

In the proposed AnoOnly, the online cluster learning through BN relies on the statistical characteristics (mean $\bm{\mu}_\mathcal{B}$ and standard deviation $\bm{\sigma}_\mathcal{B}$) within the mini-batch $\mcal{B}$. 
Guided by the law of large numbers, we hypothesize that as the batch size increases, the statistical characteristics become more representative of the distributional properties.
To investigate the effect of batch size, we perform ablation experiments by gradually increasing the batch size from 1 ($2^0$) to 128 ($2^7$) on DeepSAD \cite{ruff2019deep} integrated with AnoOnly, as shown in Fig. \ref{fig:bz}.
The results indicate that when the batch size is small (e.g., 1 or 2), the superiority of AnoOnly is suppressed due to biased statistical characteristics. 
Surprisingly, the performance gains achieved by AnoOnly tend to stabilize when the batch size is larger than 4. 
This empirical observation justifies our analysis that abundant normal data is sufficient in amount but monotonous in statistics, providing limited guidance for anomaly detection.
Furthermore, there is a negligible decline in performance as the batch size increases from 32 to 128. This is because the statistical characteristics become sufficiently reflective when the batch size reaches 32, and a larger batch size reduces the number of iterations for model weight updates.

\subsection{Anomaly score and t-SNE Visualization}
\begin{figure*}[t]
  \centering
  \includegraphics[width=0.9\linewidth]{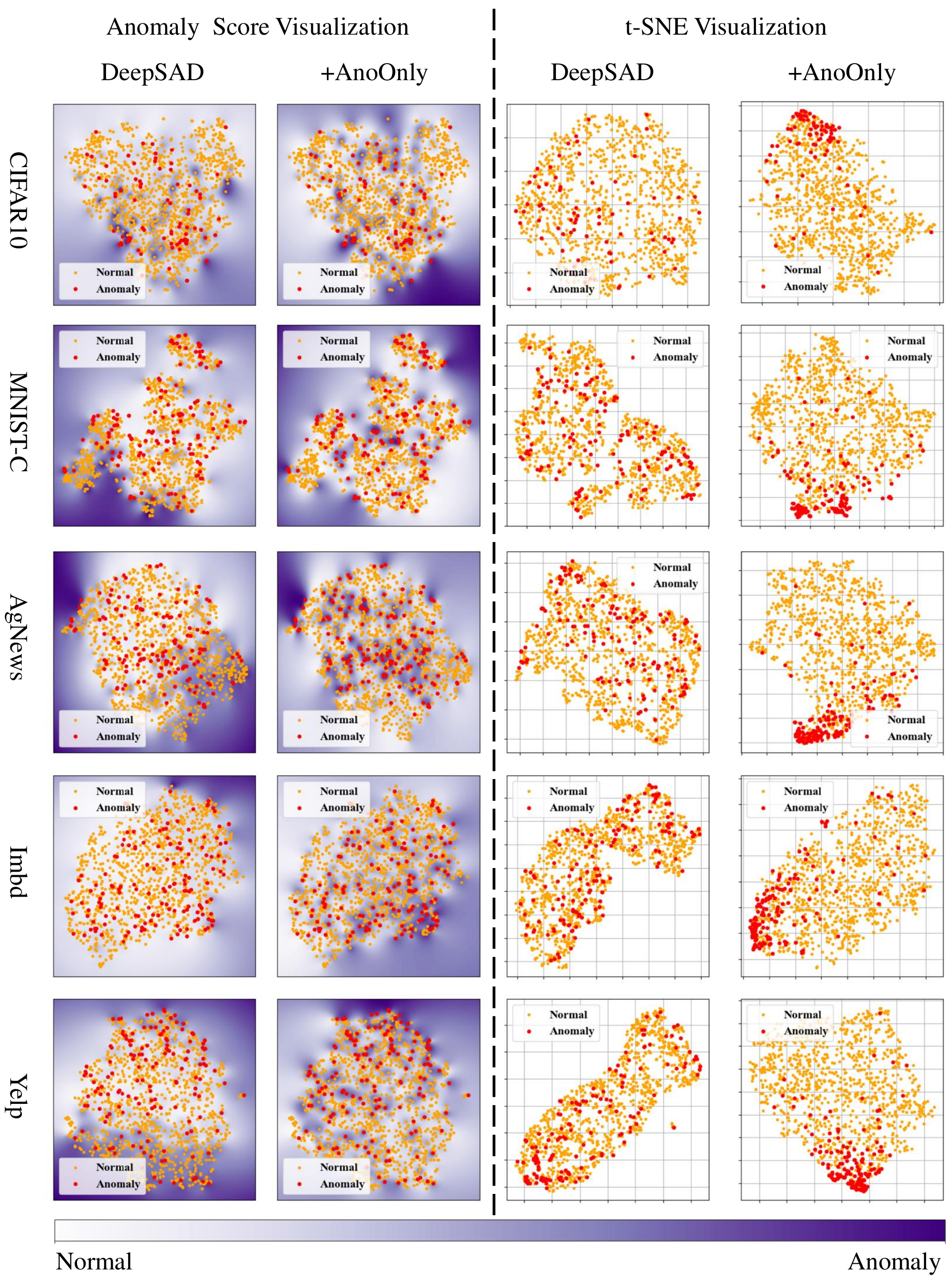}
  \caption{Visualization of predicted anomaly score on extracted features and t-SNE visualization of the hidden features before the last layer in $\mathcal{F}(\cdot; \bm{\theta})$, which are also the inputs of anomaly classifier $\mathcal{C}(\cdot; \bm{\theta}_\mcal{C})$.}
  \label{fig:viz}
\end{figure*}

\tbf{Visualization of anomaly score.}
To highlight the effectiveness of our AnoOnly framework, we provide intuitive visualizations using t-SNE \cite{van2008visualizing} to visualize the extracted features of six datasets involving diverse domains including semantic anomaly detection (\emph{CIFAR10} \cite{krizhevsky2009cifar} for CV and \emph{AgNews} \cite{zhang2015agnews} for NLP), distortion detection (\emph{MNIST-C} \cite{mu2019mnist} for CV), sentiment analysis (\emph{Imdb} \cite{maas2011imdb} for NLP), and (\emph{Yelp} \cite{he2016amazon} for NLP), in the first two columns of Fig. \ref{fig:viz}.
Then we depict the anomaly scores predicted by vanilla DeepSAD and its integration with our AnoOnly, where the darker purple color indicates higher anomaly scores.
Comparing the first column (DeepSAD) with the second column (+AnoOnly), it is evident that our AnoOnly successfully detects abnormal instances that are mixed with normal instances, which are in the center of t-SNE visualizations.
Moreover, our AnoOnly significantly enhances the discrimination between normal and abnormal data by assigning anomaly scores with a distinct gap.

\FloatBarrier

\tbf{Visualization of hidden features.} 
In addition to visualizing the extracted features using pre-trained feature extractors, we further apply t-SNE to visualize the hidden features before the last layer in $\mathcal{F}(\cdot; \bm{\theta})$, which are also the inputs of the anomaly classifier $\mathcal{C}(\cdot; \bm{\theta}_\mcal{C})$, in the last two columns of Figure \ref{fig:viz}.
Although the pre-trained feature extractors (ResNet-18 \cite{he2016resnet} for CV datasets and Bert \cite{devlin2018bert} for NLP datasets) effectively transform the raw inputs into the dense representations, the t-SNE visualizations of the extracted features in the first two columns do not exhibit clear discrimination between normal and abnormal instances.
In contrast, since the model is only supervised by the loss function specifically for anomalies in our AnoOnly, the hidden features belonging to the anomaly are clearly clustered together, leading to discriminative representations of abnormal instances.
This clustering of anomaly features serves as a solid foundation for the subsequent anomaly classifier $\mathcal{C}(\cdot; \bm{\theta}_\mcal{C})$ to effectively assign high scores to abnormal instances without misclassifying normal instances.
By visualizing the hidden features, we provide further evidence of the ability of our AnoOnly to learn discriminative representations of anomalies and improve the performance of the anomaly detector in accurately detecting abnormal instances

These visualizations provide a clear demonstration of the efficacy of our AnoOnly framework in effectively detecting anomalies and improving the discrimination between normal and abnormal instances in diverse datasets and domains.

\subsection{The generalization on unseen anomalous types}

In practical applications, anomaly detection models tend to encounter new categories of anomalies that are unseen in the limited labeled dataset $\mcal{D}_L$.
Therefore, the ability to generalize their detection capabilities to these novel classes of anomalous samples becomes a pivotal criterion in estimating the performance of anomaly detection methods.
To validate the generalization capacity of our AnoOnly framework in detecting anomalies of unseen types, we rebuild the CIFAR-10 dataset. 
In particular, we retain the real class labels of the 9 abnormal classes for the convenience of selecting anomalies from certain classes for training, and later we test the model on all 9 abnormal classes.

We train the models with the labeled dataset including 1, 3, and 5 seen classes of anomalies, respectively, and report the performance on both seen and unseen classes of anomalies in Table \ref{tab:generalization}.
Additionally, we add the ablation of the utilization of the over-sampling strategy for abnormal samples, which suffers from the generalization as mentioned above. All experiments are conducted on DeepSAD with the same labeled anomaly ratio of 10\%.

\begin{table}[h]
  \caption{Comparison of performance on seen and unseen types of anomalies (AUCROC in \%) between DeepSAD and our AnoOnly with the ablation of over-sampling.}
  \label{tab:generalization}
  \centering
  \tabcolsep=6pt
  \resizebox{0.8\linewidth}{!}{%
  {%
    \begin{tabular}{@{}lcccccc@{}}
      \toprule
      \multicolumn{1}{c}{\multirow{2}{*}{Methods}} & \multicolumn{2}{c}{1 seen} & \multicolumn{2}{c}{3 seen} & \multicolumn{2}{c}{5 seen}                                                                                                                                                                     \\ \cmidrule(lr){2-3} \cmidrule(lr){4-5} \cmidrule(lr){6-7}
      \multicolumn{1}{c}{}                               & Seen & Unseen & Seen & Unseen & Seen & Unseen                  \\ \midrule
      {DeepSAD(w/o sampling)}                         & 72.2 & 68.8 & 66.9 & 65.4 & 65.1 & 63.5           \\
      {DeepSAD(w sampling)}                         & 92.1 & \tbf{72.1} & 81.2 & 75.1 & 80.9 & 77.6           \\
      {AnoOnly(w/o sampling)}                         & 93.4 & 71.7 & \tbf{83.5} & \tbf{78.9} & \tbf{82.6} & \tbf{79.9}           \\
      {AnoOnly(w sampling)}                         & \tbf{93.6} & 71.7 & 83.1 & 76.5 & 82.3 & 78.4\\
      \bottomrule
    \end{tabular}%
  }}
\end{table}

According to Table \ref{tab:generalization}, adopting over-sampling and our AnoOnly consistently improves the performance of both seen and unseen anomalous types, which empirically demonstrates that the underfitting issue towards the anomaly learning is more crucial than the overfitting issue towards the limited abnormal instances in existing methods. We believe this insight strongly supports our motivation of rebalancing the supervision volume between normal and abnormal data.
Although the performance towards unseen anomalous types is lower than that of seen anomalous types, our AnoOnly achieves the best performance for both seen and unseen anomalous types when no fewer than 3 anomalous types are available. 
Even when selecting only 1 anomalous type for training, our AnoOnly still achieves comparable performance against the best and outperforms the vanilla DeepSAD. We conjecture the incomplete access of anomalies also helps to make the decision boundary explicit, leading to improvements of unseen anomalous types. Since our AnoOnly leverages batch normalization to capture statistical characteristics of normal data, which is more representative than the learned features supervised by prior loss functions.

In particular, when additionally adopting over-sampling in our AnoOnly, it's natural for the degradation of unseen anomalous types due to the more severe overfitting of seen anomalous types. However, over-sampling inversely worsens the AUCROC of seen anomalous types, which is contrary to the expectation.
We conjecture that over-sampling of seen abnormal instances biases the statistical characteristics captured by batch normalization to the limited seen classes, distracting the cluster learning that is crucial for the learning of normal data. 


\subsection{The robustness to label noise}
In semi-supervised anomaly detection, another chronic issue is the label noise in the unlabeled dataset $\mcal{D}_U$, where the normal data is contaminated with some abnormal data.
This data contamination is inevitable for applications in real-world scenarios \cite{jiang2022softpatch} and poses a challenge to the robustness of SSAD methods.
To estimate the robustness of our AnoOnly in the presence of label noise, we conduct experiments on the clean data setting, where we filter out the abnormal instances in the unlabeled dataset $\mcal{D}_U$. In this case, the unlabeled dataset $\mcal{D}_U$ can be regarded as the normal dataset $\mcal{D}_N$.

\begin{wrapfigure}{R}{0.48\linewidth}
  \centering
  \vspace{-5pt}
  \includegraphics[width=1.0\linewidth]{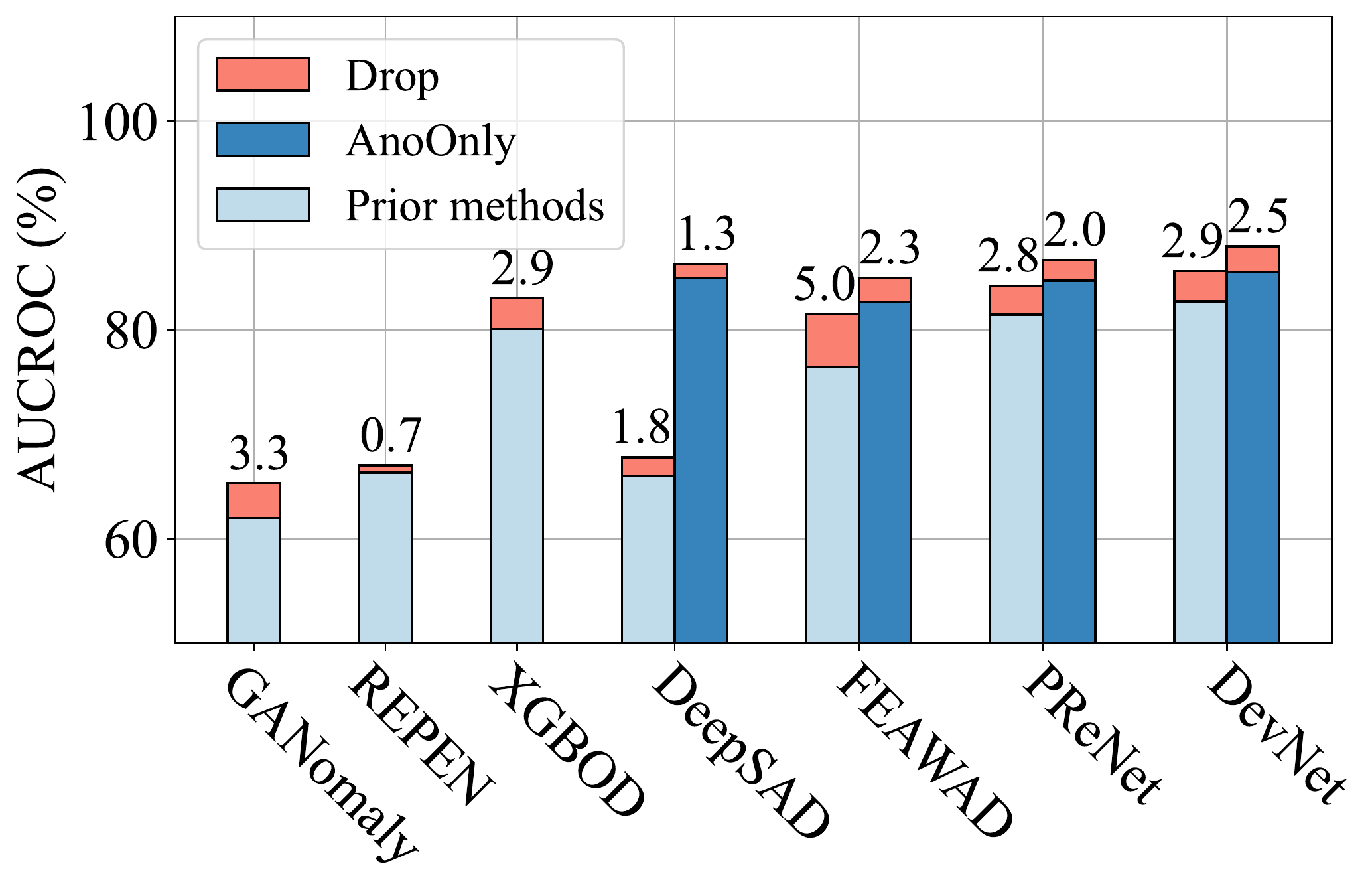}
  \caption{Overall AUCROC $(\%)$ comparison of prior SSAD methods and those integrated with our AnoOnly on ten datasets with label noise. The performance drops caused by label noise are highlighted in $\textcolor{salmon}{\blacksquare}$ and annotated in the illustration.}
  \label{fig:robust}
\end{wrapfigure}

As illustrated in Fig.\ref{fig:robust}, compared with models trained in clean datasets, prior SSAD methods exhibit obvious performance degeneration (\textcolor{red}{${\downarrow 5.0\%}$} for FEAWAD \cite{zhou2021feature} and \textcolor{red}{${\downarrow 2.9\%}$} for DevNet \cite{pang2019devnet}) when suffering from label noise.
In these SSAD methods, due to the contamination of normal data with anomalies, the unlabeled abnormal data is deemed as normal data and provides strict loss supervision $L_N$ designed for normal data during model training.
On the other hand, the model is concurrently trained by the adverse loss supervision $L_A$ on labeled anomalies.
The adversarial loss supervision on anomalies confuses the model, inducing diminished performance.

When incorporated with our AnoOnly, the performance drops of existing SSAD methods are alleviated.
Moreover, our AnoOnly enables the SSAD methods to achieve comparable performance to those trained on clean datasets, indicating enhanced robustness to label noise.
In the proposed AnoOnly, we omit the strict loss supervision $L_N$ and resort to weak supervision introduced by BN for noisy unlabeled data.
Since BN serves as implicit cluster learning to capture statistical characteristics that are less influenced by label noise, our AnoOnly is natively robust to label noise.

\section{Conclusion}
In this paper, we investigate existing semi-supervised anomaly detection (SSAD) methods and reveal that the imbalanced supervision volume derived from overwhelming normal data biases the models against anomaly perception. To redirect the model bias to anomaly detection, we propose a simple but effective framework called AnoOnly.
The proposed AnoOnly leverages batch normalization to implicitly perform online cluster learning as a form of weak supervision to replace the conventional strict loss supervision on normal data. 
When incorporated with prior methods, our AnoOnly successfully rebalances the one-sided supervision volume, impressively enhancing performance and achieving a novel SOTA. 
Furthermore, the proposed AnoOnly demonstrates strong robustness when suffering from label noise, which is practical for real-world applications.

\section*{Limitations and broader impacts}
While the proposed AnoOnly method effectively improves anomaly detection performance, the precision of detecting normal instances inevitably deteriorates due to the removal of supervision for normal data. This limitation could be problematic for some real-world scenarios such as industrial manufacture. 
This limitation also motivates research to explore a controllable trade-off of supervision between imbalanced normal and abnormal data. 
Besides, the key insight and implementation of rebalancing supervision for classes with imbalanced data distribution in our AnoOnly could be applied to other tasks such as long-tailed recognition.

\bibliographystyle{splncs04}
\bibliography{ref}

\end{document}